\documentclass[11pt,a4paper]{article}

\usepackage{enumitem}
\usepackage{graphicx}%
\usepackage{multirow}%
\usepackage{amsmath,amssymb,amsfonts}%
\usepackage{amsthm}%
\usepackage{mathrsfs}%
\usepackage[title]{appendix}%
\usepackage{xcolor}%
\usepackage{textcomp}%
\usepackage{manyfoot}%
\usepackage{booktabs}%
\usepackage{array}%
\usepackage{algorithm}%
\usepackage{algorithmicx}%
\usepackage{algpseudocode}%
\usepackage{listings}%
\usepackage{placeins}%
\usepackage{caption}%
\usepackage{array}%
\usepackage{longtable}%
\usepackage{array}%
\newcolumntype{C}[1]{>{\centering\arraybackslash}p{#1}}
\newcolumntype{L}[1]{>{\raggedright\arraybackslash}p{#1}}
\usepackage[margin=2.5cm]{geometry}
\usepackage[hidelinks]{hyperref}
\usepackage{cite}

\usepackage[final,commandnameprefix=always]{changes}

\definecolor{BrightRed}{RGB}{220, 20, 60}
\definecolor{VividOrange}{RGB}{255, 128, 0}

\setaddedmarkup{{\color{VividOrange}#1}}
\setdeletedmarkup{{\color{BrightRed}\sout{#1}}}

\definecolor{AddBlue}{RGB}{0, 102, 204}

\theoremstyle{plain}%

\theoremstyle{definition}%

\begin{document}

\title{SAMe: A Semantic Anatomy Mapping Engine for Robotic Ultrasound}

\maketitle

\vspace{-1.5em}
\begin{center}
\normalsize
Jing~Zhang$^{1,*,\dagger}$,
Duojie~Chen$^{1,2,*}$,
Wentao~Jiang$^{1}$,
Zihan~Lou$^{1}$,
Jianxin~Liu$^{3}$,
Xinwu~Cui$^{4}$,
Qinghong~Zhao$^{5}$,
Bo~Du$^{1,\dagger}$,
Christoph~F.~Dietrich$^{6}$,
Dacheng~Tao$^{7,\dagger}$

\vspace{0.6em}

\small
$^{1}$ School of Computer Science, Wuhan University, China\\
$^{2}$ Hubei Center for Applied Mathematics, Wuhan University, China\\
$^{3}$ Department of Ultrasound, The Central Hospital of Wuhan, China\\
$^{4}$ Department of Medical Ultrasound, Tongji Hospital, Tongji Medical College, Huazhong University of Science and Technology, China\\
$^{5}$ Department of Ultrasound in Medicine, Renmin Hospital of Wuhan University, China\\
$^{6}$ University Hospital, Johann-Wolfgang-Goethe University Frankfurt am Main, Germany \\
$^{7}$ College of Computing and Data Science, Nanyang Technological University, Singapore
\end{center}

\renewcommand{\thefootnote}{\fnsymbol{footnote}}
\footnotetext[1]{These authors contributed equally to this work.}
\footnotetext[2]{Correspondence to: jingzhang.cv@gmail.com; dubo@whu.edu.cn; dacheng.tao@ntu.edu.sg}

\begin{abstract}
Robotic ultrasound has advanced local image-driven control, contact regulation, and view optimization, yet current systems lack the anatomical understanding needed to determine what to scan, where to begin, and how to adapt to individual patient anatomy. These gaps make systems still reliant on expert intervention to initiate scanning. Here we present SAMe, a semantic anatomy mapping engine that provides robotic ultrasound with an explicit anatomical prior layer.
SAMe addresses scan initiation as a target-to-anatomy-to-action process: it grounds under-specified clinical complaints into structured target organs, instantiates a patient-specific anatomical representation for the grounded targets from a single external body image, and translates this representation into control-facing 6-DoF probe initialization states without any additional registration using preoperative CT or MRI.
The anatomical representation maintained by SAMe is explicit, lightweight (single-organ inference in 0.08~s), and compatible with downstream control by design.
Across semantic grounding, anatomical instantiation, and real-robot evaluation, SAMe shows strong performance across the full initialization pipeline. 
In real-robot experiments, centroid-based SAMe initialization outperformed the body-keypoint-based heuristic baseline under a budget-matched single-target setting for both liver (86.7\% versus 46.7\%) and kidney (80.0\% versus 73.3\%) initialization. 
Furthermore, The trial-level organ-hit rate reached 97.3\% for liver and 83.3\% for kidney when multiple candidate targets were available.
These results establish an explicit anatomical prior layer that addresses scan initialization and is designed to support broader downstream autonomous scanning pipelines, providing the anatomical foundation for complaint-driven, anatomically informed robotic ultrasonography.
\end{abstract}


\section{Introduction}\label{sec1}

\begin{figure*}[t]\centering
\makebox[\textwidth][c]{%
\includegraphics[width=0.97\textwidth]{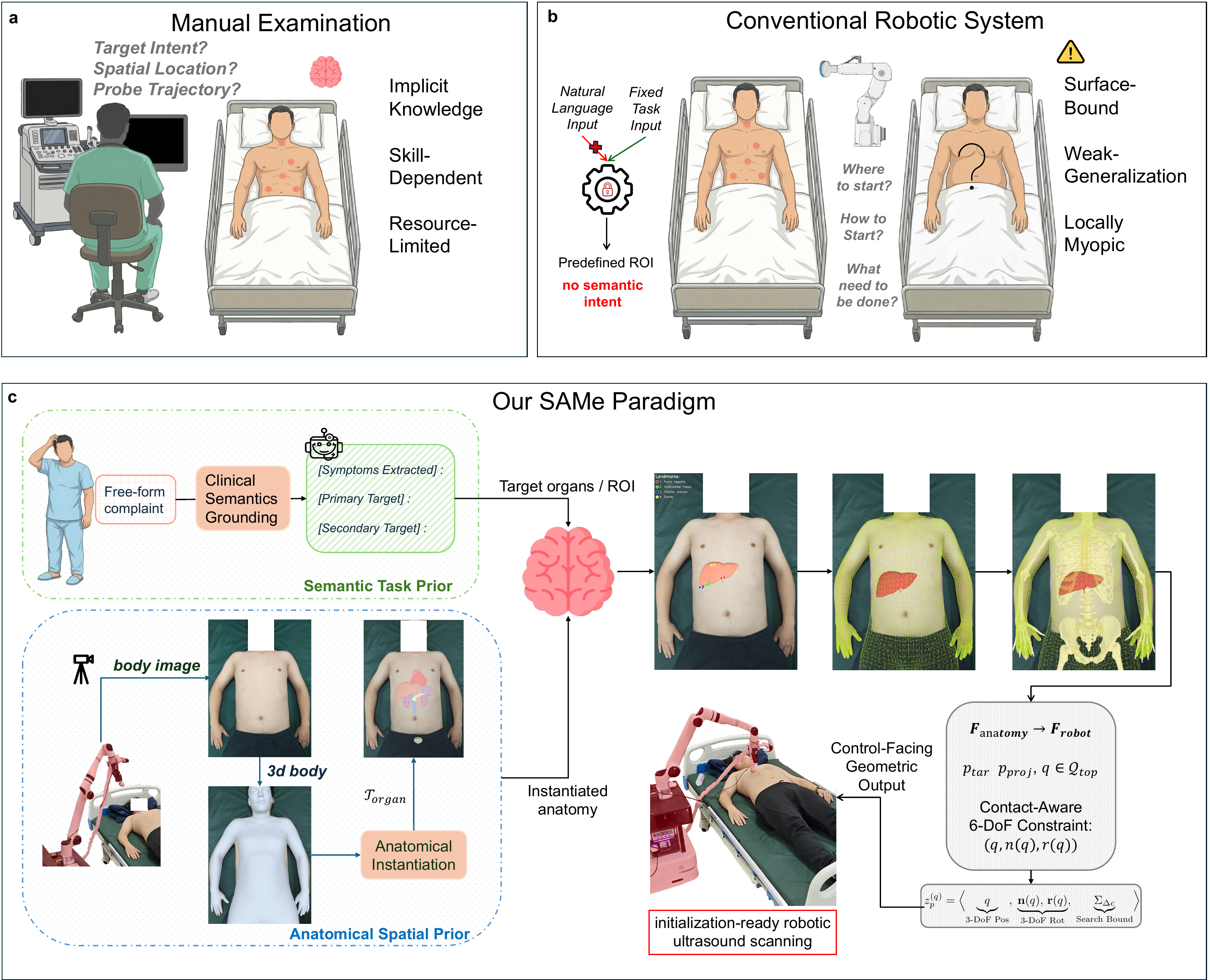}%
}
\caption{From manual expertise to autonomous robotic ultrasonography.
\textbf{(a)} Manual examination relies on implicit sonographer knowledge, making it skill-dependent and resource-limited.
\textbf{(b)} Conventional systems often rely on predefined tasks, expert targets, or body-keypoint-based heuristics, lacking patient-specific anatomical priors for scan initialization.
\textbf{(c)} SAMe bridges natural language complaints with robotic execution through a plug-in prior layer, yielding contact-aware 6-DoF initialization constraints in the form of candidate contact points, surface normals, target-directed rays, and uncertainty-aware anatomical guidance for risk-aware probe placement.}
\label{fig1}
\end{figure*}

Ultrasonography plays a central role in clinical imaging because it is real-time, radiation-free,
portable and cost-effective. It is already established across fetal~\cite{salomon2011practice,namburete2023normative}, cardiac~\cite{ulloa2021deep}, vascular~\cite{stein2008use,tahmasebpour2005sonographic}, liver~\cite{ferraioli2019ultrasound,ferraioli2018liver} and thyroid~\cite{leenhardt20132013} imaging, with organ-specific guidelines, quantitative assessment workflows and outcome-linked analyses developed in multiple domains. At the same time, ultrasound continues to expand beyond conventional bedside examination toward wearable imaging, continuous monitoring and conformal physiological sensing~\cite{lin2024fully,hu2023wearable,wang2018monitoring}. 

Expert operators can achieve efficient and reliable examinations by tightly coupling image interpretation, probe manipulation, anatomical reasoning, and clinical judgment. However, ultrasound examination remains highly dependent on manual operation and expert decision-making, with scanning strategies adapted case by case to patient-specific anatomy and acoustic conditions~\cite{beales2011reproducibility,joakimsen1997reproducibility,kojcev2017reproducibility}. As a result, high-quality acquisition still depends on trained operators, prolonged supervised training and sustained clinical support, while workforce shortages and training-capacity constraints have been reported across multiple healthcare systems~\cite{won2024sound,asrt2023staffing_survey,bls2025diagnostic_sonographers,mcgregor2020providing,sonographycanada2023hhr_submission,sor2026_ultrasound_vacancy,parker2015educating,coleman2024exploring}.

These constraints motivate robotic ultrasound and system-level automation not as a rejection of expert practice, but as a way to organize, reproduce and extend expert-quality acquisition more reliably~\cite{swerdlow2017robotic,monfaredi2015robot,huang2023review,du2024review,su2024fully,jiang2025towards}.
Within this direction, robotic ultrasound systems have made substantial progress in local image-driven control, including probe adjustment, contact-aware control, and standard-view search. Probe adjustment and task-constrained standard-view search have been demonstrated across a range of robotic settings~\cite{bi2024machine,liu2025screens,merouche2015robotic,akbari2021robot,jiang2021autonomous,wang2024autonomous,chen2023fully,jiang2023skeleton,jiang2021motion}, while more direct policy-learning approaches have also been explored for autonomous scanning behaviors~\cite{hase2020ultrasound,li2021autonomous,ning2021autonomic,bi2022vesnet,droste2020automatic,men2022multimodal,deng2021learning}.  These advances mark an important step toward partial autonomy in constrained scanning tasks~\cite{bi2024machine,liu2025screens}.  Nevertheless, most existing systems incorporate anatomical priors for local probe motion optimization~\cite{jiang2023skeleton,jiang2022towards,hennersperger2016towards,jiang2022precise}, but remain limited in their ability to determine anatomically grounded targets before scanning begins.

Taken together, current robotic ultrasound system lack a reliable anatomy-aware initialization stage before local control begins, thus still relying on expert intervention to initiate scanning. This system-level limitation has three outward manifestations, corresponding to Fig.~\ref{fig1}b. First, most existing systems are restricted to predefined tasks or require expert-specified targets, making it difficult to determine scanning targets directly from clinical intent in realistic clinical scenarios. Second, current systems lack a reliable strategy for spatial target initialization. They either depend on expert guidance, which limits autonomy, or rely on body-keypoint-based heuristics~\cite{long2024localizing,su2024fully,wang2024autonomous}, which often lead to inaccurate initialization and increase the downstream burden of corrective search and local optimization. Third, current systems also lack explicit 3D constraints imposed by internal anatomical morphology and skin-surface geometry, making the inferred probe pose or access direction still anatomically inappropriate for obtaining a valid ultrasound view. This leaves a system-level gap before local control begins: what should be scanned, where scanning should start, and how the initial probe pose should be organized, motivating the development of an explicit and unified anatomical engine.

To address these gaps, we introduce SAMe, a Semantic Anatomy Mapping Engine for robotic ultrasound.
Rather than replacing existing local controllers, SAMe is designed as a plug-in prior
layer. Its core role is to improve anatomy-aware scan initialization before local image-driven refinement begins. It addresses these limitations through three coupled components, corresponding to
Fig.~\ref{fig1}c.
Clinical Semantics Grounding converts under-specified clinical complaints or task descriptions into structured anatomical targets. Anatomical Representation Instantiation constructs a patient-specific anatomical representation of
organ location, morphology, and uncertainty in a standardized body-aligned space.
Actionable Target Initialization converts this representation into
robot-readable geometric constraints for target localization and 6-DoF probe initialization.

\begin{figure*}[t]
\centering
\includegraphics[width=\textwidth]{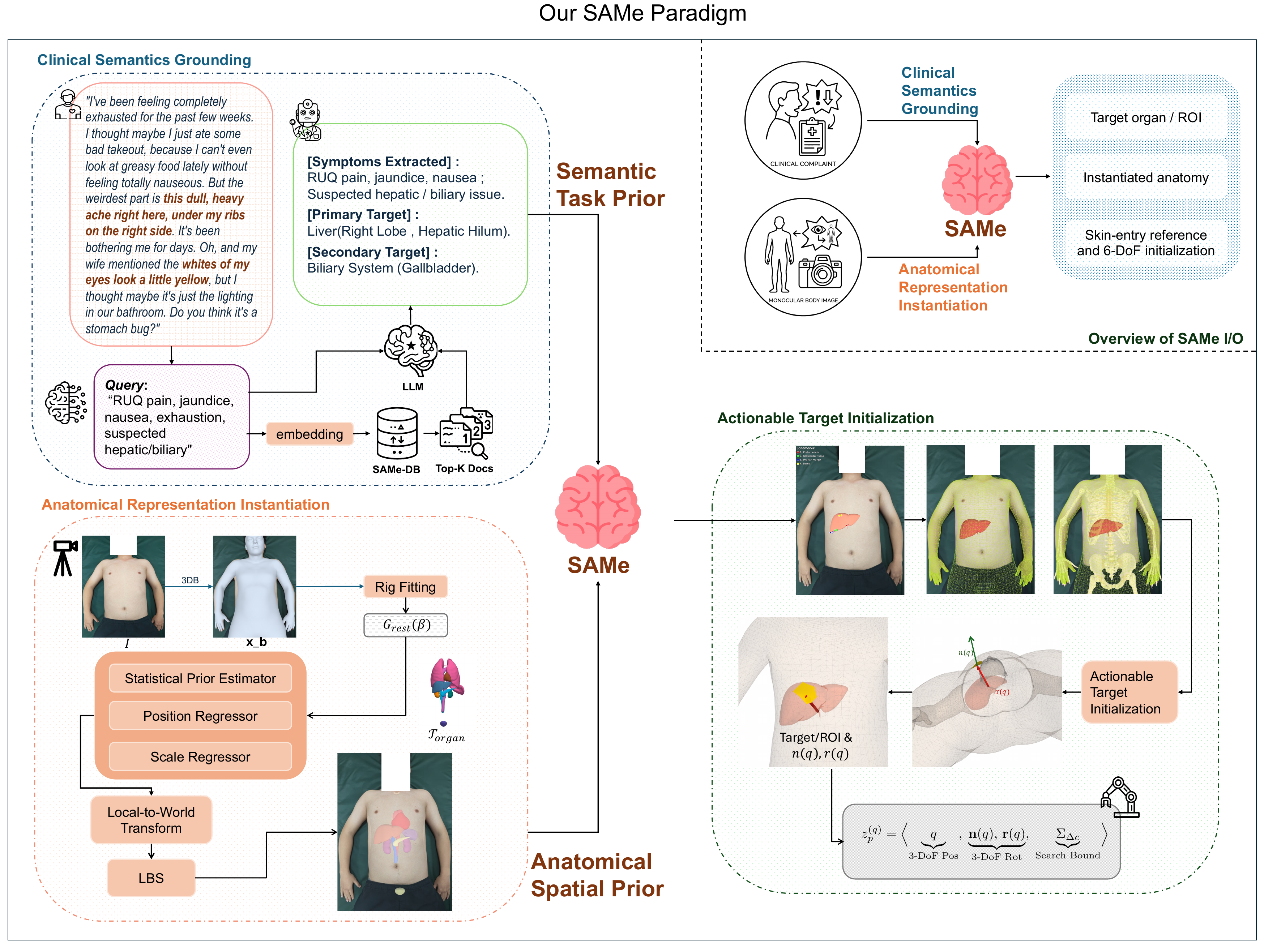}
\caption{The SAMe system architecture bridging clinical semantics to robotic execution.
\textbf{(a)} Clinical Semantics Grounding converts narrative patient complaints into structured
scanning targets through LLM-based symptom extraction and retrieval from the SAMe semantic prior
database, outputting primary and secondary anatomical targets.
\textbf{(b)} Anatomical Representation Instantiation estimates patient-specific organ location
and morphology via statistical prior estimation coupled with body surface reconstruction and
skeleton-anchored rig fitting.
\textbf{(c)} Actionable Target Initialization translates the instantiated anatomy into
robot-readable geometric cues, comprising 3-DoF contact position, 3-DoF probe orientation,
and uncertainty-bounded search regions for downstream scan initiation.}
\label{fig2}
\end{figure*}

In this study, SAMe is instantiated and evaluated in a concrete setting centered on anatomically informed scan initialization for robotic ultrasound, with semantic grounding incorporated as an upstream extension for complaint-driven target specification. The focus is placed on three coupled abilities:
1) grounding under-specified clinical complaints into structured scanning targets when semantic input is provided,
2) instantiating a patient-specific anatomical representation for the grounded target, and
3) translating this internal representation into control-facing 6-DoF initialization cues.
In this formulation, the anatomical representation is explicit and probabilistic---patient-specific in its instantiation, uncertainty-aware in its representation, and designed for posterior updating and bidirectional coupling as new observations become available.
These components are organized at the system level in Fig.~\ref{fig2}, evaluated in
Sections~\ref{sec:results-csg}--\ref{sec:results-ati}, and formalized in
Section~\ref{sec:methods}.
This instantiation is evaluated in real-robot liver and kidney localization experiments,
where SAMe improves organ-level initialization reliability and enables anatomically
meaningful target specification. These results position SAMe as a prior
layer for robotic ultrasound that addresses scan initialization and is designed to be compatible with downstream control and posterior updating, whereas full closed-loop integration and online posterior refinement remain important directions for future work.

In summary, this study makes three contributions. First, it identifies anatomy-aware initialization before local control as an important systems problem in robotic ultrasound. Second, it instantiates a prior-based initialization pathway that links patient-specific anatomical representation to control-facing 6-DoF initialization outputs, while also allowing complaint-driven semantic grounding when needed. 
Third, we validate the approach in real-robot localization experiments, showing that this prior formulation not only improves initialization reliability under a budget-matched single-target setting, but also enables multi-target anatomical coverage and anatomically meaningful target specification—capabilities beyond the reach of body-keypoint heuristics.

\section{Results}\label{sec2}

\subsection{SAMe as a system-level prior engine for robotic ultrasound}

In the present study, SAMe is instantiated as a system-level prior layer for robotic ultrasound. Its role is to organize scan initiation before local execution begins by
transforming clinical input into an explicit, task-conditioned anatomical representation and then
into robot-readable initialization outputs.

In this instantiation, SAMe operates through three coupled components. Clinical Semantics Grounding (CSG) converts under-specified clinical complaints into
structured scanning targets (Section~\ref{sec:results-csg}). Anatomical Representation
Instantiation  (ARI) builds a patient-specific anatomical representation for the grounded target in a
standardized body-aligned representation (Section~\ref{sec:results-ari}). Actionable Target
Initialization (ATI) converts this instantiated anatomy into control-facing geometric outputs, including
target-aware spatial initialization and 6-DoF probe-entry cues (Section~\ref{sec:results-ati}).

Within this formulation, the three components address three system-level questions.
CSG addresses what should be scanned. ARI addresses where the relevant anatomy is likely to be in the current subject.
ATI addresses how scanning should begin in a geometrically and
anatomically informed manner. The following sections evaluate these three capabilities in order.

\FloatBarrier
\subsection{Clinical Semantics Grounding translates narrative clinical complaints into structured scanning targets}\label{sec:results-csg}

\begin{figure*}[!t]
\centering
\includegraphics[width=\textwidth]{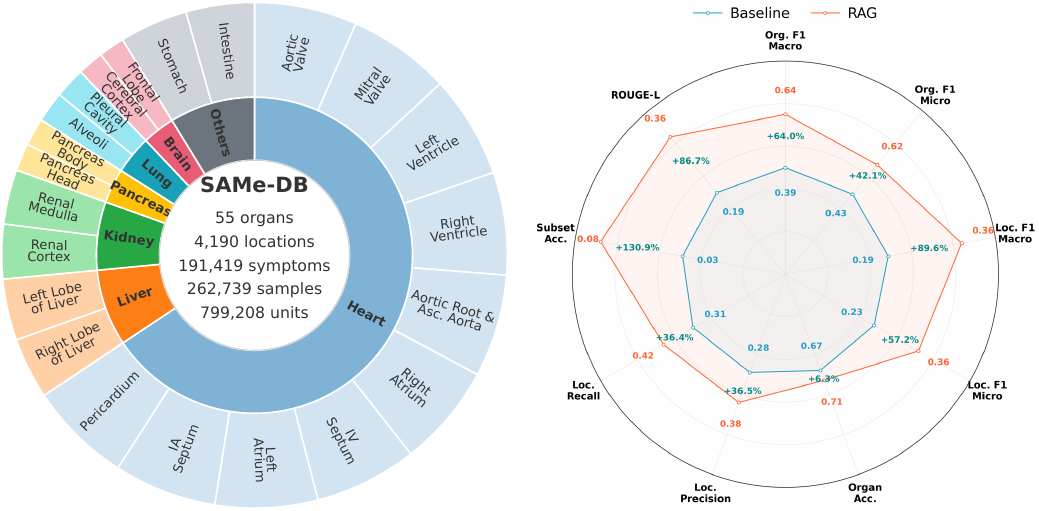}
\caption{Overview of the SAMe semantic prior database and RAG performance across LLM backends. \textbf{(a)} Sunburst visualization of the SAMe-DB ontology coverage spanning 55 organs and 4,378 anatomical locations, with 191,419 unique symptoms organized into 262,739 samples and 799,208 structured units. \textbf{(b)} Radar chart comparing RAG-enhanced versus baseline performance averaged across the evaluated LLM backends, showing consistent improvements in organ-level and location-level grounding metrics (macro and micro F1).}
\label{fig:same-db-overview}
\end{figure*}

\begin{figure*}[!t]
\centering
\includegraphics[width=\textwidth]{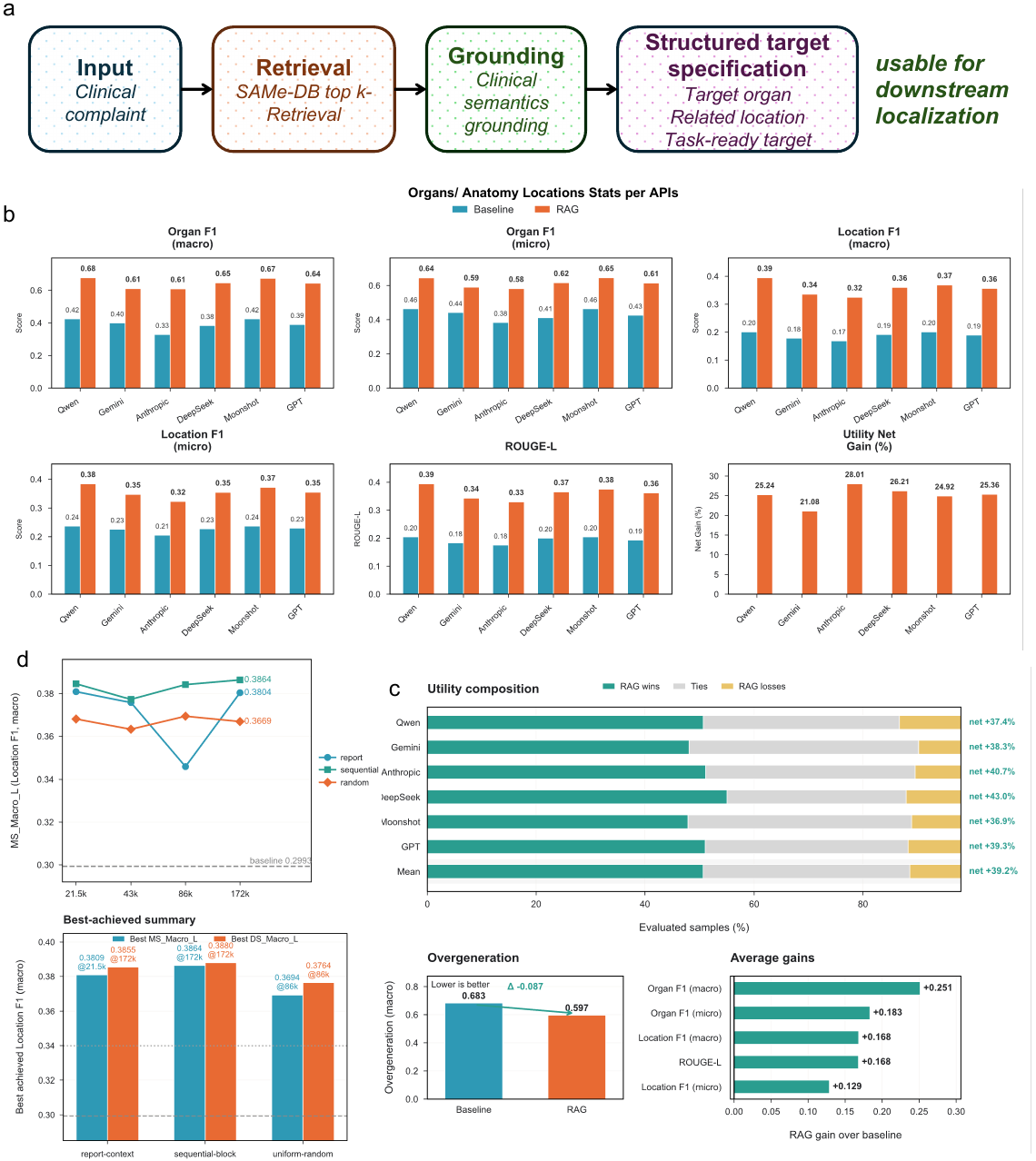}
\caption{Clinical Semantics Grounding results. \textbf{(a)} System role of the semantic layer in SAMe:
clinical complaint or report text is retrieved against the SAMe semantic prior and grounded into a
structured target specification comprising target organ, related anatomy, prioritized location or
ROI, and task-ready targets. \textbf{(b)} Baseline-versus-RAG grounding performance across evaluated model
backends. \textbf{(c)} Output-quality analysis, including pairwise utility, overgeneration, and average
gains across reported metrics. \textbf{(d)} Effects of semantic-prior construction strategy and scale, with
best-achieved summaries across the three compared backend construction policies.}
\label{fig:semantic-grounding-results}
\end{figure*}

The first system-level question addressed here is what should be scanned. SAMe can reliably ground under-specified
clinical complaints into structured scanning targets. It semantically transforms complaint or report text into a structured target specification that includes the target organ, related anatomy, prioritized location or region of interest (ROI), and task-ready targets (Fig.~\ref{fig:semantic-grounding-results}a).

To support CSG, we constructed the fine-grained RAG backend from MIMIC-IV-Note v2.2~\cite{PhysioNet-mimiciv-2.2}. As summarized in
Fig.~\ref{fig:same-db-overview}, the
resulting semantic prior database spans 55 organs and 4,378 anatomical locations, with 191,419 unique
symptoms, 262,739 total samples, and up to 799,208 structured units. The resource was organized into
structured symptom--diagnosis--organ--anatomical-location units that match the output form required
by downstream target specification, providing structured clinical language and anatomical targets for following layer. Detailed construction procedures are provided in
Section~\ref{sec:methods-semantic}.

At the system level, this resource serves as the task-aligned semantic prior backend for CSG in SAMe. SAMe uses this indexed semantic prior as external evidence for grounding
under-specified clinical input into structured, anatomically actionable targets. We test whether this semantic prior backend is useful in deployment settings. Specifically, we examine a scenario where complaint or report text is provided to an LLM backend, and evaluate whether retrieval from the SAMe semantic prior improves final target grounding. To keep this stage
lightweight and easy to deploy, retrieval was implemented with a simple Faiss-based index~\cite{johnson2019billion} that can
be built and queried on a single GPU.

Within this plug-in setting, the first question was how the semantic-prior backend should be
constructed so that it remains both useful and practical. We therefore first examined three backend
construction policies across four matched scales to test whether semantic grounding quality depends
only on data volume or also on how the retrieval evidence is organized. As shown in
Fig.~\ref{fig:semantic-grounding-results}d, construction strategy had a clear and systematic effect
on both retrieval quality and final grounding quality. Increasing scale generally improved
performance, but scale alone did not explain the trend: the sequential-block strategy, which
preserves local report structure, showed the strongest best-achieved location grounding and the
most stable behavior across scales, whereas the uniform-random strategy consistently
underperformed. This pattern suggests that semantic grounding benefits from clinically coherent
evidence organization rather than from a larger but less structured retrieval pool. The 172k
sequential-block semantic-prior backend was therefore selected for subsequent evaluation because it
provided the best overall balance between grounding performance and practical backend scale.

With the semantic-prior backend fixed, the key question was whether under-specified clinical
complaints could be grounded reliably enough to serve as SAMe's front-end semantic prior layer
across different model backends. As shown in Fig.~\ref{fig:semantic-grounding-results}b, the gains
were consistent across the evaluated backends, indicating that the benefit was not tied to
any particular model. Importantly, the improvement was especially pronounced at the
anatomical-location level, where grounded outputs become more directly actionable for downstream
anatomical instantiation and target initialization. On a random held-out test set of 1,000 symptom
descriptions excluded from the retrieval resource, adding the SAMe semantic prior increased
location-level F1 from 0.188 to 0.357 in macro averaging and from 0.227 to 0.356 in micro
averaging, while organ-level F1 improved by 0.251 in macro averaging and 0.183 in micro
averaging\footnote{Macro F1 denotes the unweighted mean of per-class F1 scores over organs or anatomical locations, whereas micro F1 is computed by pooling prediction outcomes across all samples and classes before calculating F1.}. Backend-specific maxima varied by model: Qwen achieved the strongest absolute
anatomy-level grounding after RAG, whereas the largest organ-level gains were observed for
Anthropic in macro F1 (+0.280) and DeepSeek in micro F1 (+0.204). Together, these results show
that clinical text can be grounded into structured organ- and anatomy-level targets with
substantially greater reliability when supported by the semantic prior.

The semantic prior changed the character of the generated target specifications. As shown in
Fig.~\ref{fig:semantic-grounding-results}c, RAG-enhanced outputs were preferred in pairwise
utility evaluation, achieving a mean win rate of 50.6\% against 11.4\% losses and yielding a net
gain of 39.2\%. At the same time, overgeneration was reduced from 0.683 to 0.597, and the
average gain remained positive across all reported metric families. 
Together, these results establish CSG as the front-end semantic prior layer of SAMe rather than a generic retriever. The grounded output consists of target organs, anatomy-level locations or ROIs, and task-ready cues, matching the structured input required by ARI for subsequent anatomical instantiation. In the real-robot experiments below, liver and kidney targets were fixed 
since the body-keypoint baseline does not support anatomy-aware target specification, a paired CSG comparison was not feasible to isolate the downstream anatomy-to-action evaluation, so CSG is not used to claim a direct improvement in organ-hit rate.

\subsection{Anatomical Representation Instantiation ensures reliable localization across body variations}\label{sec:results-ari}

Having established what should be scanned, the next system-level question is where the relevant
anatomy is likely to be. Downstream
initialization depends on a representation that estimates 
where the relevant organ lies and how far it extends under the subject's current body configuration.
ARI therefore aims to recover the patient-specific anatomical
hypothesis needed for initialization, rather than relying on fixed templates or weakly conditioned
surface priors.

We tested whether SAMe could recover organ placement and support more reliably than surface-driven or weakly conditioned alternatives when body habitus changes across subjects. Evaluation was performed on the Quadra-HC dataset~\cite{gutschmayer2025whole,quadrahc_zenodo2025}, using 35 held-out cases, 11 organs, and 385 case-organ pairs in total. SAMe was compared against four baselines for localization performance: no personalization (Template mean), a generic statistical organ-shape prior without subject-specific anchoring (PCA subspace; inspired by statistical body--organ shape models such as BOSS~\cite{shetty2023boss}), weak external conditioning from overall body habitus alone (Global body-shape-only), and surface-only conditioning from the observed torso geometry (Skin-only conditioning) (Table~\ref{tab:ari_baselines}).

\begin{table*}[!t]
\centering
\begin{minipage}[t]{0.24\textwidth}
\vspace{0pt}
\centering
\textbf{(a)}\par\vspace{0.35em}
\includegraphics[width=0.9\linewidth]{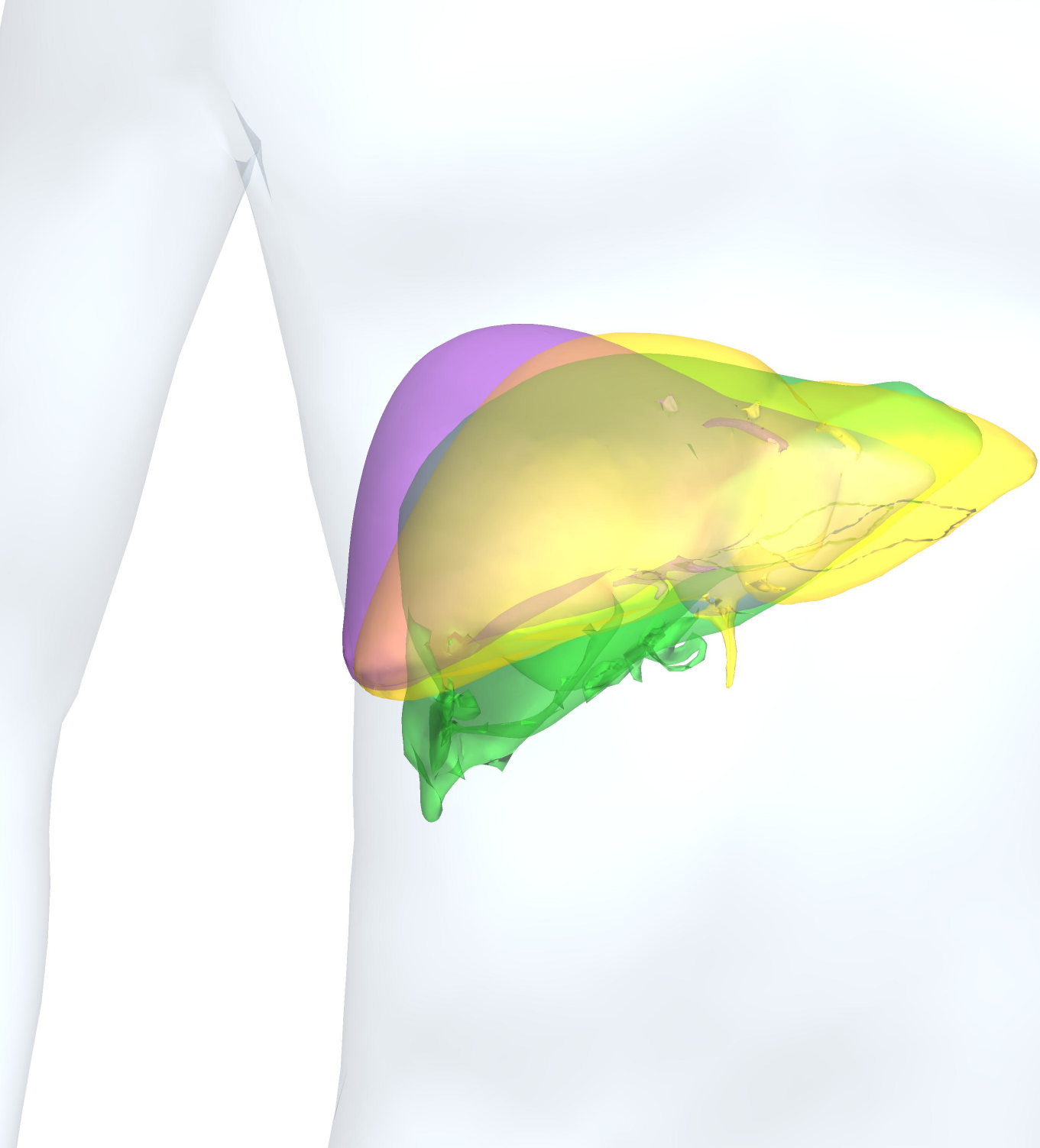}
\end{minipage}
\hfill
\begin{minipage}[t]{0.735\textwidth}
\vspace{0pt}
\centering
\textbf{(b)}\par\vspace{0.35em}
{\scriptsize
\setlength{\tabcolsep}{5pt}
\renewcommand{\arraystretch}{1.08}
\resizebox{\linewidth}{!}{%
\begin{tabular}{@{}l c >{\centering\arraybackslash}p{1.05cm} >{\centering\arraybackslash}p{1.05cm} >{\centering\arraybackslash}p{1.05cm} c c@{}}
\toprule
Baseline & \shortstack{Centroid\\err. $\downarrow$\\(mm)} & \multicolumn{3}{c}{\shortstack{Per-axis centroid err. $\downarrow$\\(mm)}} & \shortstack{Scale\\err. $\downarrow$\\(\%)} & \shortstack{Support\\IoU $\uparrow$} \\
\cmidrule(lr){3-5}
 &  & x & y & z &  &  \\
\midrule
Template mean & 32.97 & \textbf{8.72} & 13.61 & 25.08 & 19.37 & 0.291 \\
PCA subspace & 38.17 & 15.90 & 18.72 & 21.71 & 21.08 & 0.231 \\
Global body-shape-only & 59.55 & 18.63 & 21.65 & 46.49 & 22.45 & 0.179 \\
Skin-only conditioning & 50.45 & 18.97 & 21.91 & 32.69 & 18.95 & 0.215 \\
SAMe & \textbf{22.55} & 9.02 & \textbf{9.23} & \textbf{15.09} & \textbf{16.10} & \textbf{0.391} \\
\bottomrule
\end{tabular}
}
}
\end{minipage}
\caption{Anatomical Representation Instantiation results. \textbf{(a)} Liver instantiation
example. Green denotes ground-truth liver volume, yellow denotes the SAMe prediction, and purple
denotes the Template mean baseline prediction. \textbf{(b)} Quantitative comparison across the
current baseline models on the Quadra-HC evaluation subset. Entries are shown as mean values over
35 cases $\times$ 11 organs = 385 case-organ pairs. Centroid error denotes mean Euclidean centroid
error. Per-axis centroid error denotes the mean absolute centroid error along the x, y, and z
axes. Scale error denotes bounding-box extent relative error, and Support IoU denotes
bounding-box-support IoU.}
\label{tab:ari_baselines}
\end{table*}

As shown in Table~\ref{tab:ari_baselines}, SAMe achieved the best overall localization accuracy among all compared methods, reducing mean centroid error to 22.55~mm versus 32.97--59.55~mm for the baselines. Relative to the strongest baseline by overall centroid error, this corresponds to a 31.6\% reduction. This result indicates that patient-specific instantiation provides stronger localization than global habitus cues or external surface observations alone.

The advantage pronounced along the z axis further clarifies where this gain comes from. SAMe achieved per-axis centroid errors of 9.02 / 9.23 / 15.09~mm along x / y / z. Compared with the weakly conditioned baselines, errors were reduced substantially along all axes, with the most pronounced advantage along z, where the Global body-shape-only and Skin-only conditioning baselines showed large deviations (46.49 and 32.69~mm, respectively). This pattern is consistent with the intended role of rig-anchored conditioning: it regularizes organ position around a population average, and adapts organ placement to subject-specific anatomical configuration.

 SAMe improved not only point localization but also the recovery
  of organ spatial support. It achieved the lowest scale error (16.10\%) and
  the highest Support IoU (0.391), improving over the strongest baseline by
  15.0\% in scale error and 34.4\% in Support IoU. This shows that the
  instantiated anatomy is better aligned both in where organs are placed and how far they extend in space. This distinction matters for
  downstream robotic use, because scan initialization benefits from knowing
  organ extent and occupancy.

  This advantage is enabled by the explicit low-dimensional representation
  used by SAMe. The structured parametric model (MHR~\cite{ferguson2025mhr}) with explicit skeletal structure and predicts an anatomically interpretable state keeping the representation compact, uncertainty-aware, and
  directly usable for robot-facing initialization. In our implementation, once
  the subject-specific MHR representation is available, full-organ prior
  inference runs in as little as 0.76~s on CPU, and liver-only inference in
  0.08~s, highlighting its lightweight character and practical suitability for
  deployment. Organ-wise results are provided in the Supplementary 
  Table~\ref{tab:supp_ari_organs}.

To provide a complementary, anatomy-specific view of instantiation quality, we further evaluated on the liver 
using two practical criteria. The target-in-organ inclusion rate at 10~mm (83.3\%) counts each internal target prediction that is either inside the liver or falls outside by no more than 10~mm from the liver surface. As illustrated in Table~\ref{tab:ari_baselines}a, this perspective emphasizes that predicted targets land in anatomically usable regions for real scan initiation. 

Together, these results establish ARI as more than a localization module: it recovers the patient-specific anatomical representation needed before robot-facing initialization, with sufficient spatial fidelity and anatomical usability to support the stage examined next.

\subsection{Actionable Target Initialization translates instantiated anatomy into improved real-robot probe initialization}\label{sec:results-ati}

\begin{figure*}[!t]
\centering
\includegraphics[width=\textwidth,height=0.88\textheight,keepaspectratio]{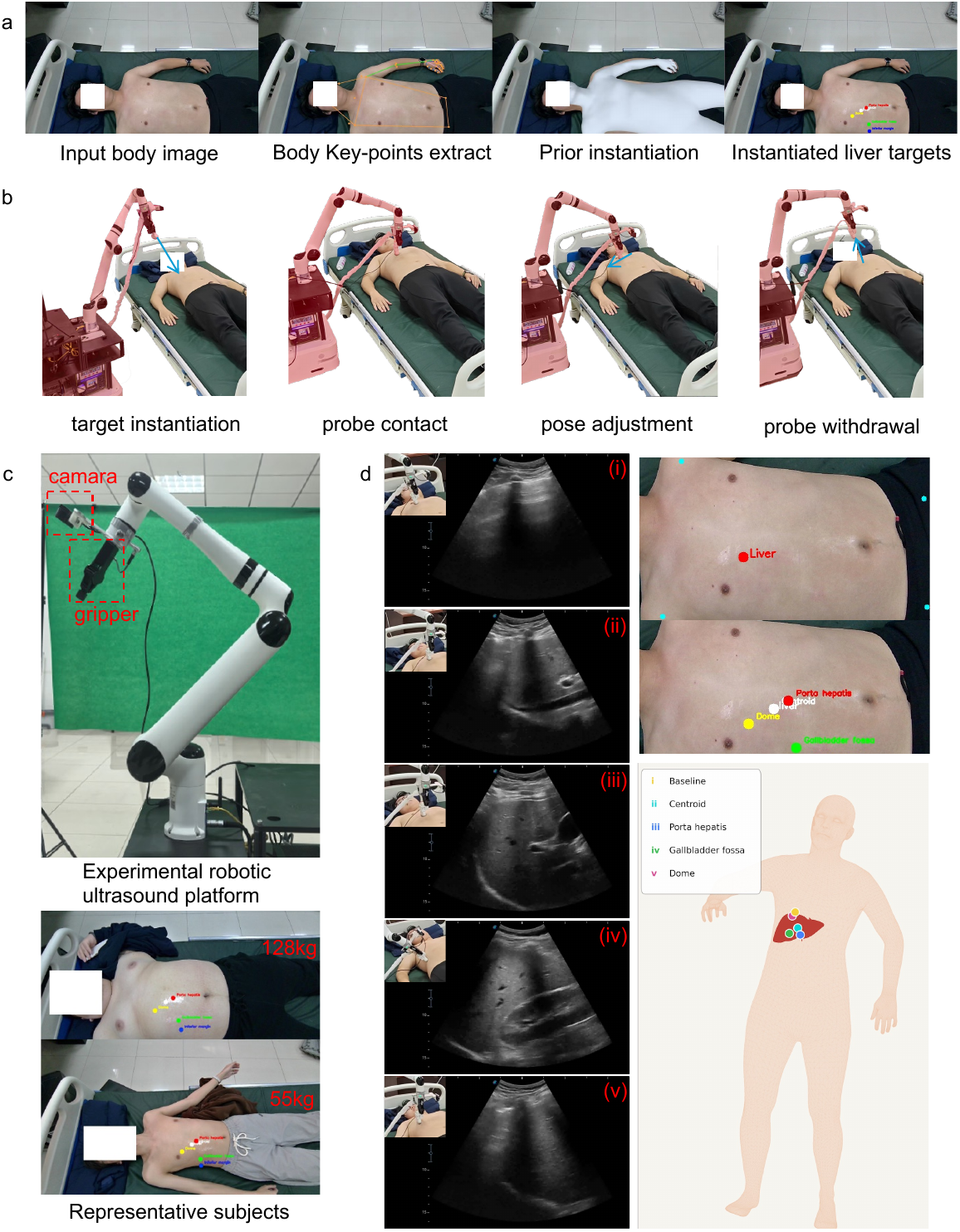}
\caption{Actionable Target Initialization in real robotic ultrasound. \textbf{(a)} Front-end
instantiation from an external body image to patient-specific liver targets. \textbf{(b)} Robotic
execution at an instantiated target. \textbf{(c)} Experimental platform and representative
subjects spanning substantial body-habitus variation. \textbf{(d)} Ultrasound returns from
instantiated liver targets, establishing point-to-view correspondence between instantiated target,
executed probe placement, and resulting ultrasound observation.}
\label{fig:ati-overview}
\end{figure*}

Having shown that grounded anatomy can be instantiated in a patient-specific body frame, the next
system-level question is how to initialize scanning for robotics. At this
stage, the issue is not anatomical validity alone, but whether the internal anatomical targets can
be converted into better robot-facing initialization. Knowing where anatomy is likely to lie is
not sufficient to determine probe contact, entry direction, or whether the initial view will
support downstream local optimization. ATI addresses this question by
converting instantiated anatomy into control-facing geometric outputs for robot-facing
initialization, while remaining complementary to image-driven refinement.

To test this robot-facing utility, we compared the full SAMe initialization stack with
body-keypoint-based heuristics. These heuristics follow the
external-geometry-based initialization strategies used in prior robotic ultrasound
pipelines~\cite{long2024localizing,su2024fully}. It first estimated body keypoints from a
monocular body image with ViTPose~\cite{xu2022vitpose}. It then predicted coarse liver and
right-kidney target points using a geometric interpolation model calibrated from more than 200 body
images annotated by expert clinicians. The local surface normal at the predicted contact point was
used as the default probe direction. By contrast, SAMe used a
single RGB image and SAM 3D Body~\cite{yang2026sam} to build a subject-specific MHR-aligned body
representation~\cite{ferguson2025mhr}, instantiate target-related internal points from the
learned anatomical prior, and convert them into target-conditioned entry geometry. This
comparison therefore evaluates the system-level effect of the complete
initialization formulation. In the present experiments, four anatomy-relevant internal targets were defined for the liver and four for the kidney. 
For each anatomy-relevant target listed in Table~\ref{tab:liver_kidney_accuracy}, the robot-facing output consisted of candidate skin-contact locations, target-directed approach vectors, and contact-aware 6-DoF initialization poses.

The full robot-facing chain is summarized in Fig.~\ref{fig:ati-overview}. Panel a shows the
conversion from a single external body image to patient-specific internal liver targets; panel b
shows robotic execution at one instantiated target; panel c situates the method in the real
platform across subjects with substantial body-habitus variation; and panel d shows the initial
ultrasound views returned from instantiated liver targets. Together, these panels establish a
practical point-to-view correspondence between designated internal targets, executed probe
placement, and first ultrasound observations.
The experimental platform shown in Fig.~\ref{fig:ati-overview}c comprised an RML63-B 6-DoF robotic arm (RealMan Intelligent Technology, Beijing, China) fitted with an EG2-SF16 electric gripper, a Gemini~335 stereo depth camera (Orbbec Inc., Shenzhen, China) for external body observation, and a SonoScape E11 portable ultrasound system with a C1-6A single-crystal convex array probe (SonoScape Medical Corp., Shenzhen, China) for image acquisition.

The side-entry example in Fig.~\ref{fig:ati-overview}d(iv) shows that SAMe can initialize the probe from lateral or angled directions, rather than only from the front body surface.
This matters because
effective initialization depends on probe--skin contact and acoustic coupling, not only geometric
proximity to the target.

We therefore evaluated whether this anatomy-aware initialization improved scan onset in real robotic ultrasound under the practical conditions.
The human-participant evaluation was conducted under institutional ethics approval and written informed consent, as detailed in Section~\ref{sec:methods}. The study included healthy young male volunteers who were not used for training the anatomical prior or for selecting the evaluated initialization targets. Specifically, the evaluation included 25 subjects for liver scanning and 20 subjects for kidney scanning, covering a broad body-habitus range with BMI from 15.8 to 35.5 and body weight from approximately 55 to 128~kg. Because multiple anatomy-relevant initialization targets were evaluated for each subject, the final analysis included 75 liver initialization trials and 60 kidney initialization trials. These trial-level results summarize all valid initialization attempts, whereas subject-level summaries were used to assess stability across participants with shared body habitus, organ depth, acoustic windows, and respiratory conditions.

All acquired ultrasound images were reviewed by expert clinicians. For both methods, organ-level initialization success was defined as locating the intended organ or target region while obtaining a visible, clinically usable initial view suitable for subsequent scanning and local optimization. For SAMe, which predicts anatomy-specific target points, an additional anatomy-level criterion tested whether the acquired image was anatomically consistent with the designated target point. Binary success was assigned only when all applicable criteria were satisfied.

\begin{table*}[!t]
\centering
\small
\setlength{\tabcolsep}{5pt}
\renewcommand{\arraystretch}{1.05}

\begin{tabular}{@{}llcccc@{}}
\toprule
& & \multicolumn{1}{c}{Trial-level organ-hit rate} & \multicolumn{2}{c}{Subject-level organ-hit rate} & \\
\cmidrule(lr){3-3}\cmidrule(lr){4-5}
Organ & Target & $n/N$ (\%) & Mean $\pm$ SD & 95\% CI & Anatomy match \\
\midrule
\multirow{6}{*}{Liver}
& \shortstack{Baseline} & 35/75 (46.7\%) & 46.7 $\pm$ 33.3 & [32.9, 60.4] & -- \\
& Centroid & 65/75 (86.7\%) & 86.7 $\pm$ 27.2 & [75.4, 98.0] & \textbf{50/75 (66.7\%)} \\
& Porta hepatis & 66/75 (88.0\%) & 88.0 $\pm$ 23.3 & [78.4, 97.6] & 25/75 (33.3\%) \\
& Gallbladder fossa & 66/75 (88.0\%) & 88.0 $\pm$ 25.2 & [77.6, 98.4] & 7/75 (9.3\%) \\
& Dome & 35/75 (46.7\%) & 46.7 $\pm$ 38.5 & [30.8, 62.6] & 16/75 (21.3\%) \\
& \textbf{SAMe Hit} & \textbf{73/75 (97.3\%)} & \textbf{97.3 $\pm$ 13.3} & \textbf{[91.8, 100.0]} & \textbf{--} \\
\midrule
\multirow{6}{*}{Kidney}
& \shortstack{Baseline} & 44/60 (73.3\%) & 73.3 $\pm$ 35.2 & [56.9, 89.8] & -- \\
& Centroid & 48/60 (80.0\%) & 80.0 $\pm$ 34.9 & [63.7, 96.3] & \textbf{43/60 (71.7\%)} \\
& Renal hilum & 47/60 (78.3\%) & 78.3 $\pm$ 34.7 & [62.1, 94.6] & 34/60 (56.7\%) \\
& Upper pole & 39/60 (65.0\%) & 65.0 $\pm$ 41.1 & [45.7, 84.3] & 31/60 (51.7\%) \\
& Lower pole & 47/60 (78.3\%) & 78.3 $\pm$ 32.9 & [62.9, 93.7] & \textbf{43/60 (71.7\%)} \\
& \textbf{SAMe Hit} & \textbf{50/60 (83.3\%)} & \textbf{83.3 $\pm$ 33.3} & \textbf{[67.7, 98.9]} & \textbf{--} \\
\bottomrule
\end{tabular}

\caption{Actionable Target Initialization results. Real-robot initialization outcomes for body-keypoint-based heuristics and SAMe anatomy-aware targets. Organ hit denotes clinically usable organ-level initialization. Trial-level results report the proportion of successful scans across all valid initialization attempts (75 liver trials, 60 kidney trials). Subject-level summaries report the mean $\pm$ standard deviation of per-subject success rates across independent participants ($n = 25$ liver, $n = 20$ kidney), with 95\% confidence intervals. Anatomy match denotes consistency between the acquired image and the designated anatomical target point. SAMe Hit denotes organ-level success when any of the four SAMe anatomy-aware targets yields a usable initialization and is reported as a multi-target coverage metric.}
\label{tab:liver_kidney_accuracy}
\end{table*}

At the trial level, centroid-based SAMe initialization substantially outperformed the body-keypoint baseline for both liver (86.7\% vs.\ 46.7\%) and kidney (80.0\% vs.\ 73.3\%); allowing any anatomy-aware candidate to count as a hit (\textbf{SAMe Hit}) further raised coverage to 97.3\% and 83.3\%, respectively (Table~\ref{tab:liver_kidney_accuracy}). Beyond organ access, SAMe uniquely supports anatomy-level matching, which body-keypoint heuristics cannot assess: centroid targeting achieved the highest liver anatomy-match rate (66.7\%), whereas peripheral or boundary-sensitive landmarks such as the dome and gallbladder fossa were harder to match; kidney targets showed a narrower match range (51.7--71.7\%). Together, these results indicate that anatomy-aware initialization improves not only where the robot aims and the probability of a usable first view, but also the capacity to resolve anatomical subregions within the scanned organ.

To verify that these trial-level gains reflect stable cross-subject behaviour rather than a few favourable subjects, we computed per-subject success rates. For liver initialization, the 95\% confidence intervals of SAMe centroid (75.4--98.0\%) and the baseline (32.9--60.4\%) were non-overlapping, and liver SAMe Hit yielded the most stable subject-level performance (95\% CI: 91.8--100.0\%; Table~\ref{tab:liver_kidney_accuracy}). The strong liver result carries direct clinical relevance: the liver is the standard initial target in FAST protocols~\cite{scalea1999focused}, a primary acoustic window for right renal imaging~\cite{koratala2019point}, and the most common site for detecting hemoperitoneum in abdominal trauma~\cite{rozycki1998early}, making automated liver initialization directly relevant to emergency ultrasonography. Kidney initialization showed larger subject-level dispersion than liver, reflecting the deeper anatomical location and stronger dependence on acoustic window quality; nevertheless, SAMe-based targeting consistently exceeded the baseline at the subject level.  Taken together, the trial-level and subject-level analyses provide complementary evidence: the former quantifies per-attempt reliability, while the latter confirms that the observed advantage of anatomy-aware initialization generalizes across subjects rather than arising from a small number of favourable cases. 

To isolate the contribution of ATI beyond anatomical instantiation alone, we performed a small
paired ablation on held-out subjects, with details provided in the
\hyperref[supp:methods]{Supplementary Methods}. The full SAMe stack was compared with an ARI-only
variant using the same instantiated centroid targets but without the contact-aware ATI stage. ATI
improved centroid-based organ hit and anatomy match for both liver and kidney. These
results support ATI as a control-facing stage that improves the practical executability of
instantiated anatomical targets, although the effect remains modest in this small ablation.

To examine whether the main real-robot findings were consistent across body habitus, we performed
a BMI-stratified analysis on the 20 subjects who completed both the liver and kidney experiments.
These subjects included the full kidney group and a shared subset of the liver group. This allowed
liver and kidney initialization to be examined within the same participant group. The resulting
patterns are summarized in Table~\ref{tab:bmi_boundary}.

\begin{table}[!t]
\raggedright
\begin{minipage}[t]{\textwidth}
\vspace{0pt}
\centering
\scriptsize
\textbf{(a)}\par\vspace{0.3em}
\setlength{\tabcolsep}{2.5pt}
\renewcommand{\arraystretch}{1.08}
\resizebox{\linewidth}{!}{%
\begin{tabular}{@{}ll*{8}{c}@{}}
\toprule
& & \multicolumn{2}{c}{\begin{tabular}[t]{@{}c@{}}Low BMI\\($<20$)\\2 subjects\end{tabular}} &
\multicolumn{2}{c}{\begin{tabular}[t]{@{}c@{}}Middle Low\\BMI $[20,25)$\\11 subjects\end{tabular}} &
\multicolumn{2}{c}{\begin{tabular}[t]{@{}c@{}}Middle High\\BMI $[25,30)$\\5 subjects\end{tabular}} &
\multicolumn{2}{c}{\begin{tabular}[t]{@{}c@{}}High BMI\\($\geq 30$)\\2 subjects\end{tabular}} \\
\cmidrule(lr){3-4}\cmidrule(lr){5-6}\cmidrule(lr){7-8}\cmidrule(l){9-10}
Organ & Target & O & A & O & A & O & A & O & A \\
\midrule
\multirow{6}{*}{Liver}
& \shortstack{Baseline} & 33.3\% & -- & 42.4\% & -- & 40.0\% & -- & 50.0\% & -- \\
& Centroid & 100.0\% & 66.7\% & 84.8\% & 66.7\% & 86.7\% & 80.0\% & 50.0\% & 50.0\% \\
& Porta hepatis & 100.0\% & 0.0\% & 90.9\% & 51.5\% & 86.7\% & 53.3\% & 50.0\% & 0.0\% \\
& Gallbladder fossa & 100.0\% & 16.7\% & 97.0\% & 12.1\% & 80.0\% & 6.7\% & 50.0\% & 16.7\% \\
& Dome & 66.7\% & 33.3\% & 51.5\% & 33.3\% & 26.7\% & 6.7\% & 66.7\% & 16.7\% \\
& \textbf{SAMe Hit} & \textbf{100.0\%} & \textbf{--} & \textbf{100.0\%} & \textbf{--} & \textbf{100.0\%} & \textbf{--} & \textbf{66.7\%} & \textbf{--} \\
\midrule
\multirow{6}{*}{Kidney}
& \shortstack{Baseline} & 83.3\% & -- & 84.8\% & -- & 53.3\% & -- & 50.0\% & -- \\
& Centroid & 100.0\% & 66.7\% & 87.9\% & 81.8\% & 60.0\% & 60.0\% & 66.7\% & 50.0\% \\
& Renal hilum & 100.0\% & 83.3\% & 87.9\% & 63.6\% & 60.0\% & 40.0\% & 50.0\% & 33.3\% \\
& Upper pole & 100.0\% & 33.3\% & 69.7\% & 63.6\% & 53.3\% & 40.0\% & 33.3\% & 33.3\% \\
& Lower pole & 100.0\% & 100.0\% & 87.9\% & 75.8\% & 53.3\% & 53.3\% & 66.7\% & 66.7\% \\
& \textbf{SAMe Hit} & \textbf{100.0\%} & \textbf{--} & \textbf{87.9\%} & \textbf{--} & \textbf{66.7\%} & \textbf{--} & \textbf{83.3\%} & \textbf{--} \\
\bottomrule
\end{tabular}
}
\par\vspace{1em}
\textbf{(b)}\par\vspace{0.3em}
\includegraphics[width=\linewidth]{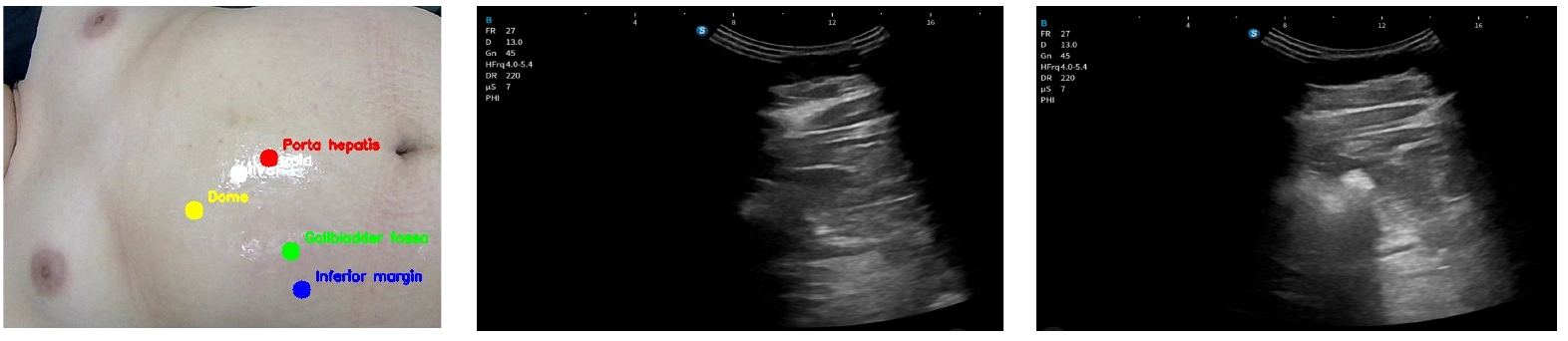}
\caption{\textbf{(a)} BMI-stratified subgroup analysis on the shared participant group. Values denote success rates. O denotes organ-level access, and A denotes anatomy-level target matching. Body-keypoint-based heuristics do not produce anatomy-specific target matches. \textbf{(b)} Failure case (BMI 35.5), showing a superior offset in the predicted initialization region. Uncontrolled respiration likely further increased the anatomy--probe mismatch.}
\label{tab:bmi_boundary}
\end{minipage}
\end{table}

The BMI-stratified results were broadly consistent with the main findings in Table~\ref{tab:liver_kidney_accuracy}. In subjects with BMI below 30, SAMe maintained its advantage across a broad range of body habitus rather than only in a narrow average-body setting. The low-BMI subjects further showed that the anatomy-aware prior generalized to lean body configurations, where reduced soft-tissue thickness and more prominent skeletal landmarks could otherwise challenge surface-driven estimation. This pattern suggests that anatomy-aware initialization remains useful under body-habitus variation, while extreme body habitus can expose the limits of a one-shot static prior. The main failure pattern is examined below.

Although overall organ-level hit rates were high, broader failure patterns
remained informative. Two trials were complete
failures in which both body-keypoint-based heuristics and all four SAMe targets yielded no usable
view, both originating from the same subject (BMI 35.5).
Retrospective review showed that the predicted initialization region was biased
superiorly, causing the initial probe contacts to land over bowel gas rather than
the usable liver acoustic window. This mismatch likely reflects two concurrent
factors: extreme body habitus pushing the external observation beyond the
learned prior distribution, and uncontrolled deep breathing introducing
respiration-dependent liver motion not represented in the static, one-shot prior.
More broadly, this failure highlights a fundamental boundary of the current
formulation: SAMe provides a one-shot, image-conditioned anatomical hypothesis
without online feedback. 

Large initial estimation errors may still require feedback-driven correction during execution, for example through ultrasound image-based verification or force--contact monitoring. Integrating such real-time feedback into the initialization loop remains an important direction for future work.

Taken together, these results establish ATI as the control-facing
output layer of SAMe rather than a simple geometric post-processing step. Fig.~\ref{fig:ati-overview}
shows that the instantiated anatomy can be carried through the full robot-facing chain from body
image to executed probe placement and returned ultrasound view, while
Table~\ref{tab:liver_kidney_accuracy} shows that this translation improves practical scan on set and
reveals where anatomy-specific targets are easier or harder to realize. In this role, SAMe does
not merely estimate anatomy; it converts instantiated anatomy into control-facing geometric outputs
for robot-facing initialization, improving both where and how robotic ultrasound begins.

\FloatBarrier

\section{Discussion}\label{sec12}

Ultrasound is indispensable in clinical care, yet access to high-quality ultrasound examination 
remains limited by the shortage of skilled operators. Robotic ultrasound is therefore compelling 
not merely as an automation technology, but as a potential route toward scalable, standardized, 
and clinically accessible imaging. However, as also reflected in recent work toward expert-level 
autonomous ultrasonography, including large-scale imitation learning~\cite{jiang2025towards}, much 
of the current progress in robotic ultrasound has focused on the ``hand'' and ``eye'' of the 
system, including local image-driven control, contact regulation, and view optimization. By 
contrast, the ``brain''-level functions responsible for task understanding, anatomical 
interpretation, and action organization remain underdeveloped. This is the system-level gap that 
SAMe is intended to address.

A useful way to understand this gap is to ask why expert manual ultrasound works so well in the 
first place. Skilled operators do not follow a fixed motor routine; they interpret symptoms and 
examination goals to decide which organs or substructures deserve attention. They also maintain a 
rough but actionable prior over where those targets are likely to lie in the current patient. 
That expectation is adapted to body shape, pose, and surface cues rather than relying only on 
fixed geometric relations between sparse landmarks. When the current view is incomplete or 
uninformative, they update this internal model and redirect the search accordingly. They also 
reason about feasibility, for example by anticipating poor acoustic windows or anatomically 
inaccessible directions caused by ribs, shadows, or gas. In other words, expert scanning depends 
on an internal semantic--anatomical model that is continuously updated during the examination. 
SAMe formalizes this previously implicit layer as an explicit computational component, not as a replacement for 
downstream control, but as a structured engine for grounding task intent, instantiating 
anatomical hypotheses, and organizing action-facing initialization.

More broadly, SAMe provides a patient-anchored anatomical substrate for future robotic-ultrasound digital twins. Tudor et al.~\cite{tudor2025scoping} point out, following the NASEM formulation~\cite{willcox2023foundational}, that a medical digital twin is expected to be personalized, dynamically updated, predictive, and decision-relevant, with bidirectional interaction between the virtual and physical systems treated as central to the concept. At the same time, they also show that only a small minority of published healthcare systems currently satisfy these criteria, while many are more accurately described as personalized digital models or digital shadows. Drummond and Gonsard~\cite{drummond2024definitions} likewise note that much of the current medical literature falls short of a fully coupled twin and therefore argue for a more practical patient-digital-twin definition centered on multidimensional, patient-specific information that informs decisions. Against this background, the value of SAMe is not that it already realizes a complete digital twin for robotic ultrasound, but that it contributes one of the key ingredients such a twin would require: an explicit patient-specific anatomical representation. In SAMe, organ placement and extent are represented in a compact body-anchored form rather than entangled with dense surface geometry alone, and the prior is formulated probabilistically so that uncertainty is preserved rather than hidden. This makes the representation patient-bound, individually instantiated, and update-compatible by design: it is tied to a specific subject, differs across subjects rather than collapsing to a generic template, and is structured so that later observations can revise the state rather than replace it. In this sense, SAMe establishes a solid anatomical foundation for future robotic-ultrasound digital twins and may, by strengthening that twin substrate, also support later world-model-style developments built on top of it.

SAMe encodes this state compactly---organ placement, extent, orientation cues, surface-contact candidates, and uncertainty in a shared frame---while keeping the anatomical reasoning separate from downstream control so that the representation remains interpretable and reusable across scanning stages. Practically, the design is lightweight and plug-in by construction. It requires only a retrieval-based semantic stage and a single monocular body image, without preoperative CT/MRI or additional registration sweeps. Full-organ prior inference runs in 0.76~s on a desktop CPU, with liver-only inference in 0.08~s, reflecting the efficiency of the low-dimensional parameterization compared with denser occupancy-style formulations~\cite{kats2026depth,henrich2025looc}. This prior layer can also complement recent large-scale learning approaches~\cite{jiang2025towards} by supplying anatomical representation before learning-based scanning begins.

The present work has three limitations that also point to our next steps. First, the present study did not include direct
comparison with recent depth-based external-to-internal localization
methods~\cite{henrich2025looc,kats2026depth}, which operate under substantially different sensing and output
assumptions from the monocular-RGB setting studied here. Second, SAMe has not yet been integrated
into a complete scanning workflow, so its value beyond initialization has not been tested in a
closed-loop setting; we are currently extending the system toward this full scanning workflow in
simulation, including integration with recent learning-based robotic ultrasound methods. Third,
although the state interface is already explicit and structured for revision, an online update
mechanism has not been implemented: incoming ultrasound observations cannot yet revise the
anatomical belief during scanning, and we are now developing a Bayesian revision loop that updates
this belief in the same probabilistic state representation. These are gaps in implementation, not in
system philosophy. The present work validates the anatomical representation substrate through semantic
grounding, patient-specific instantiation, and real-robot initialization; online posterior
revision and full closed-loop integration are natural extensions that the formulation is designed
to accommodate.

\section{Methods}\label{sec11}\label{sec:methods}

In this section, we describe the implementation and evaluation of SAMe, including semantic-prior construction, canonical anatomical representation learning, skeleton-conditioned organ prior estimation, actionable target initialization, real-robot ultrasound evaluation, and statistical analysis. The full system workflow is summarized in Fig.~\ref{fig2}.

Human participants were involved only in the real-robot ultrasound evaluation; semantic-prior construction and anatomical-prior learning used public or controlled-access datasets described below. All human-participant experiments for real-robot ultrasound evaluation were approved by 
The Biomedical Ethics Committee of Wuhan University (approval no.WHU-LFMD-IRB2026027). Written informed consent was obtained from all participants before participation, after the study procedures, robotic ultrasound protocol, ultrasound data collection, and research use of anonymized data had been explained. The real-robot ultrasound experiments were conducted only for method evaluation and were not used for clinical diagnosis or treatment decision-making. Participant-derived data were de-identified before analysis, and no personally identifiable information is released. All acquired ultrasound images used for outcome assessment were reviewed by qualified clinicians. Written consent was obtained for publication of anonymized participant-derived visual materials shown in the article or Supplementary Information.

\subsection{System overview and problem formulation}

To support more intelligent robotic ultrasound in clinically realistic workflows, SAMe is
formulated as a plug-in system-level prior engine that links clinical intent, patient-specific
anatomy, and robot-readable initialization. SAMe is designed to
augment existing scanning pipelines rather than replace their local control modules, including
image-driven probe adjustment, force-aware contact control, and standard-view optimization. 

A central design requirement of SAMe is practical deployment under clinically realistic conditions,
without additional preoperative CT/MRI, extra ultrasound sweeps for registration, or external
point-cloud reconstruction. Instead, anatomically grounded initialization is derived from
information that is already available before, or at the start of, scanning: a clinical intent
variable $c$, capturing the symptom description and examination objective, and a single monocular
image $I$. The image is processed by an external body-estimation procedure to obtain body
observations $x_b$, including body-shape coefficients, rig or skeletal parameters, and
body-surface geometry. Together, $c$ and $x_b$ support the instantiation of a patient-specific
anatomical representation $z_a$, which serves as the explicit prior for downstream initialization
before local image-based refinement begins.

As introduced in Section~\ref{sec1}, SAMe comprises three coupled components: Clinical
Semantics Grounding (CSG), Anatomical Representation Instantiation (ARI), and Actionable
Target Initialization (ATI). CSG maps the clinical intent $c$ to a grounded target organ $o$
and a prioritized anatomical region or ROI $r$. The task type $\tau$ (for example,
localization, screening, or standard-view acquisition) is inferred at grounding time by the
LLM from the query together with retrieved semantic evidence rather than stored as an explicit
field in the retrieval backend. ARI combines these task-level targets with the body
observations $x_b$ derived from the monocular image to produce a patient-specific anatomical
representation $z_a$. ATI then uses $z_a$ together with body-surface cues $x_s$, where
$x_s \subseteq x_b$ includes, in particular, skin-surface normals, to derive an abstract
initialization output $z_p$ comprising ROI candidates, contact-aware 6-DoF probe
initialization poses, and tangency constraints.

Clinical Semantics Grounding maps clinical intent to a grounded target specification for robotic
ultrasound and can be written as
\begin{equation}
\label{eq:same-semantic-prior}
(o,\tau,r) = f_{\mathrm{CSG}}(c).
\end{equation}
Here, $c$ denotes the clinical complaint or task description, $o$ the grounded target organ,
$\tau$ the task type, and $r$ the prioritized anatomical region or ROI.
In the present implementation, the semantic-prior backend explicitly supports grounding of organ
and anatomy-level targets, whereas $\tau$ is inferred by the LLM from the query and retrieved
evidence rather than looked up as an explicit backend field. This semantic prior therefore
specifies what should be examined, which task should be prioritized, and where initialization
should begin; for downstream use, the highest-ranked combination is taken to guide anatomically
informed probe initialization and subsequent local refinement.

Anatomical Representation Instantiation combines the grounded task specification with the body
observations to generate a patient-specific anatomical representation, described by
\begin{equation}
\label{eq:same-anatomy-instantiation}
z_a = f_{\mathrm{ARI}}(o,\tau,x_b).
\end{equation}
Here, $x_b$ denotes body observations derived from external body sensing, including skeletal,
shape, and surface cues, and $z_a$ denotes the resulting patient-specific anatomical representation.
This hypothesis encodes explicit multi-organ geometric hypotheses, including organ location, scale,
orientation cues, and uncertainty in a shared body-aligned representation. In the present
framework, this representation is realized through a rig-anchored anatomical model that separates
organ placement from morphology, while preserving template-level anatomical semantics, including
meshes and landmarks, for downstream robotic initialization.

Actionable Target Initialization derives geometric initialization for robotic scanning from the
instantiated anatomy, the task type, and body-surface cues. This stage is described by
\begin{equation}
\label{eq:same-actionable-init}
z_p = f_{\mathrm{ATI}}(z_a,\tau,x_s).
\end{equation}
Here, $x_s \subseteq x_b$ denotes the subset of body observations used specifically for
initialization, particularly surface cues such as skin normals, and $z_p$ denotes the abstract
initialization output of this stage. In the present implementation, this abstract output is
instantiated as an organ-specific explicit control-facing geometric state, denoted by $\zeta_k$.
These quantities translate anatomical hypotheses into
probe-relevant geometric constraints that can be used to initialize downstream local control and
image-based refinement.

The present study constructs and validates these three stages as a prior framework for clinical
semantics grounding and anatomically informed initialization in robotic ultrasound. Online
observation-conditioned updating and fully autonomous closed-loop scanning are beyond the scope of the current
work.

\subsection{Data}\label{sec:methods-data}

Two data sources were used in this study, aligned with the first two SAMe components: (i) whole-body
CT volumes with tissue segmentations for Anatomical Representation Instantiation, and (ii) a
clinical-text corpus for Clinical Semantics Grounding.
\subsubsection{Anatomical imaging data for Anatomical Representation Instantiation (CT)}\label{sec:methods-data-ct}

For Anatomical Representation Instantiation, the Healthy-Total-Body-CTs collection hosted
by The Cancer Imaging Archive (TCIA), which provides low-dose whole-body CT scans from 30 healthy
adult research participants together with tissue segmentations \cite{selfridge2023healthy}, was
used. Skin, skeletal, and organ surfaces were reconstructed from the provided labels, and the
current anatomical asset set for the prior-learning pipeline comprised 11 ultrasound-relevant
anatomical structures: aorta, bladder, heart, left kidney, right kidney, liver, lung, pancreas,
spleen, thyroid, and inferior vena cava.Focusing on healthy whole-body
data reduces pathology-specific deformation and preserves a consistent outer-body reference for
inside-from-outside anatomical modeling. These CT segmentation labels were converted into
subject-specific surface meshes using the Marching Cubes algorithm~\cite{lorensen1998marching},
which serve as the geometric input to the canonical anatomical representation pipeline in
Section~\ref{sec:methods-canonical}. Surface reconstruction and preprocessing details are
provided in the Supplementary Methods under \hyperref[supp:ct_recon]{CT surface reconstruction and
preprocessing}.
Unless otherwise noted, this TCIA cohort was used for canonical anatomical representation
construction and prior learning, whereas ARI localization performance was evaluated as a
cross-dataset generalization test on 35 held-out Quadra-HC cases, with all compared baselines run
on the same evaluation set.

\subsubsection{Clinical text corpus for Clinical Semantics Grounding}\label{sec:methods-data-text}

For Clinical Semantics Grounding, the retrieval resource and all reported offline experiments were
based on MIMIC-IV-Note v2.2, a controlled-access collection of deidentified free-text clinical
notes hosted on PhysioNet~\cite{PhysioNet-mimiciv-2.2}. The data were accessed after
obtaining PhysioNet credentialed access, completing the required CITI Data or Specimens Only
Research training, and signing the applicable PhysioNet Data Use Agreement.

In the final study, MIMIC-IV-Note was used only as an offline resource for semantic grounding,
including symptom-to-organ and symptom-to-anatomy support, structured semantic abstraction, and
retrieval-augmented task-type inference. The resulting system was used only for method
development and offline experiments, not for clinical decision-making, diagnosis, treatment
recommendation, or patient management.

All processing was performed locally and offline, with no re-identification attempts and no
release of raw note excerpts or patient-level text in the manuscript or external artifacts. The
final-study corpus should therefore be
interpreted as a method-development resource for semantic grounding rather than evidence of
deployment-ready clinical performance. Details of text cleaning, chunking, embedding-based
indexing, and semantic-prior generation are provided in Section~\ref{sec:methods-semantic} and the
Supplementary Methods.

\subsection{Semantic prior distillation, indexing, and task grounding}\label{sec:methods-semantic}

For Clinical Semantics Grounding, a task-aligned semantic prior is constructed from the final-study
MIMIC clinical-text corpus defined in Section~\ref{sec:methods-data-text}. The goal of this layer
is to ground clinical intent into structured anatomical targets that are directly useful for
robotic ultrasound initialization.

Each distilled record is represented as a structured semantic unit
$u=\left(s,d,o,a,b\right)$, where $s$ denotes the symptom or sign description, $d$ the diagnosis, $o$ the standardized target organ, $a$ the anatomy-level target locations, and $b$ the
supporting textual basis. This representation preserves both clinical traceability and anatomical
usefulness for downstream grounding. Task type is not stored as an explicit field in these
retrieval records; instead, it is inferred at grounding time by the LLM from the clinical query
and the retrieved semantic evidence.

To obtain these units, a staged distillation pipeline is used. First, symptom and sign evidence is
extracted from the raw clinical text while excluding diagnosis labels and treatment decisions.
Second, organ anchors are derived from physician-authored diagnostic sections to preserve medically
grounded organ assignment. Third, each symptom--organ pair is mapped to anatomy-level targets,
yielding structured semantic units that can be indexed as anatomy-aware retrieval records. Prompt
templates, normalization rules, and implementation details are provided in the Supplementary
Methods under \hyperref[supp:semantic_layer]{Semantic layer: structured unit definition, staged
distillation, and retrieval records}.

At inference time, a clinical query or symptom description is used to retrieve semantically
relevant records from the indexed semantic-prior backend, and the retrieved evidence is aggregated
into a grounded target specification, represented in the present formulation by
$(o,\tau,r)$: the grounded target organ $o$, a prioritized anatomical region or ROI $r$ resolved
from the retrieved anatomy-level locations $a$, and a task type $\tau$ inferred by the LLM from
the query together with the retrieved evidence. Related secondary structures may also be retained
as auxiliary context. In the present implementation, the distilled records are
embedded and indexed with a Faiss-based vector index in a lightweight retrieval setting that can be
built and queried on a single GPU. Because retrieval operates directly on the
symptom--diagnosis--organ--anatomical-location representation, this layer functions as a
task-aligned semantic prior backend for zero-shot grounding from clinical descriptions to
ultrasound-relevant anatomical targets, while still allowing task type to be inferred at query
time rather than prescribed by the backend itself.

\subsection{Canonical anatomical representation and prior learning}\label{sec:methods-canonical}

\begin{figure*}[t]
\centering
\includegraphics[width=\textwidth]{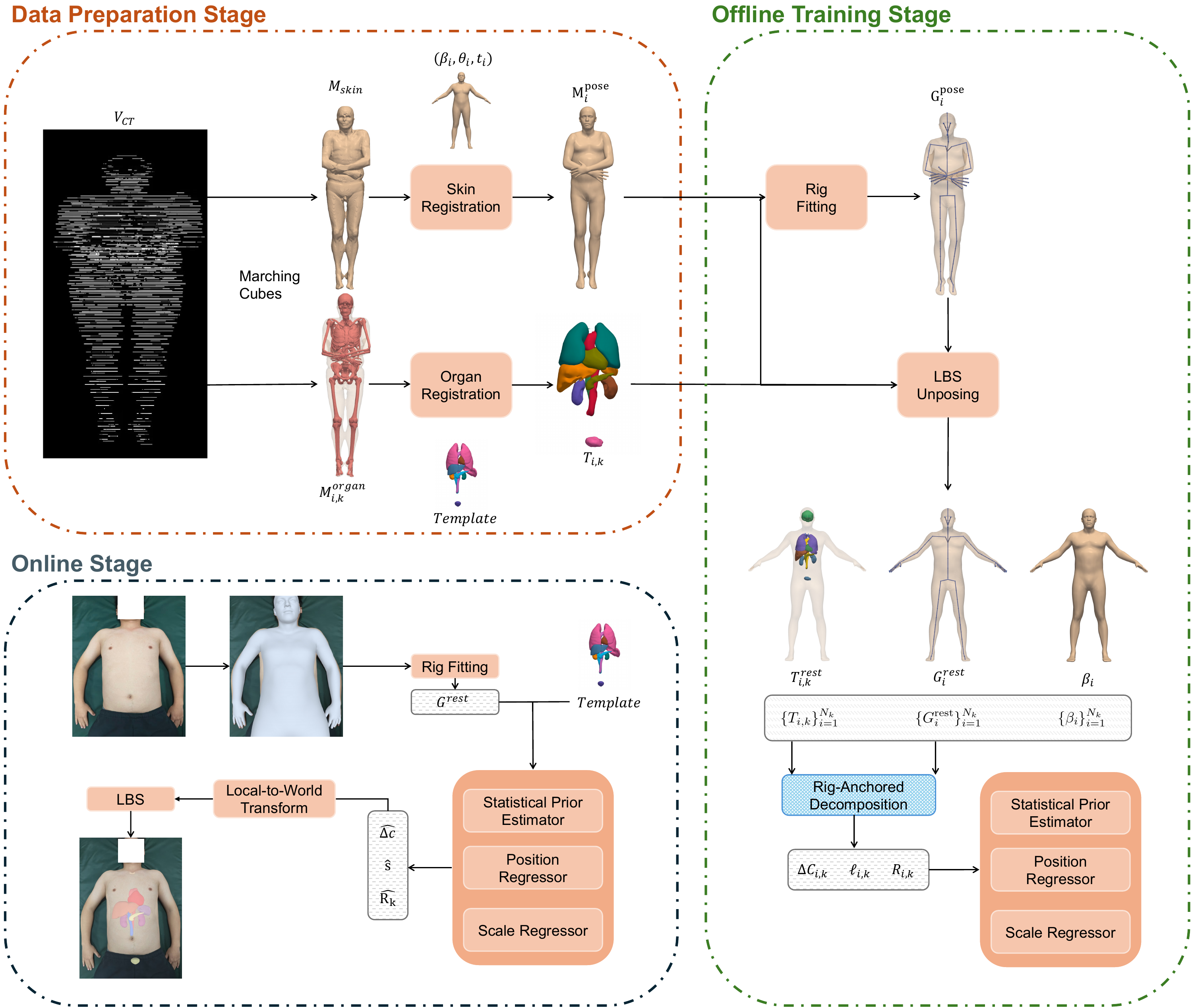}
\caption{Overview of the organ-layer modeling pipeline. Starting from CT-derived skin, skeletal,
and organ meshes, skin-based SMPL-H registration is performed, the fitted body state is converted to the
MHR rig, subject-specific organ anatomy is unposed/canonicalized into the MHR rest-pose space, and the
rig-consistent organ-layer representation is obtained for downstream prior learning and online
instantiation.}
\label{fig3}
\end{figure*}

The full organ-layer registration and canonicalization workflow is summarized in Fig.~\ref{fig3}.

To support Anatomical Representation Instantiation, a canonical anatomical representation is first constructed
in which organ geometry can be compared, parameterized, and learned consistently
across subjects. Raw CT-derived anatomy is not directly suitable for this purpose, because organ
meshes are observed in subject-specific poses and do not share a common topology. A body-aligned representation is therefore
built that removes pose-induced variability while preserving
patient-specific anatomical context. This canonical representation serves as the geometric basis
for subsequent prior learning, in which organ placement, scale, and uncertainty are estimated in a
form directly usable for downstream robotic initialization. Supporting implementation details for
mesh assets, explicit rig-transform definitions, and registration procedures are provided in the \hyperref[supp:methods]{Supplementary Methods}.

\subsubsection{Canonical representation construction}\label{sec:methods-canonical-construction}

Starting from CT-derived skin and organ meshes, a registered body frame is first established that
provides a rig-consistent anatomical reference for each subject. Specifically, an SMPL-H
body model~\cite{SMPL-X:2019,MANO:SIGGRAPHASIA:2017,SMPL:2015} is fit to the skin surface and geometry-derived 3D keypoints, and the fitted body
state is then converted to the MHR rig~\cite{ferguson2025mhr,MHR:2025}, which provides the explicit skeletal semantics required for subsequent
internal organ anchoring and synthesis stages. In the main text, the two core objectives
of this registration step, namely bidirectional skin-surface alignment and keypoint consistency, are retained:
\begin{equation}
\label{eq:skin-registration-objective}
\mathcal{E} =
\lambda_{\mathrm{surf}}\bigl(\mathcal{E}_{\mathrm{s2m}} + \mathcal{E}_{\mathrm{m2s}}\bigr)
+ \lambda_{\mathrm{j3d}} \sum_m w_m \left\| J_m(\beta_i,\theta_i,t_i) - p_m \right\|_2^2.
\end{equation}
Here, $\mathcal{E}_{\mathrm{s2m}}$ and $\mathcal{E}_{\mathrm{m2s}}$ denote the bidirectional
surface-alignment terms, $J_m(\beta_i,\theta_i,t_i)$ the $m$th model joint under shape, pose, and
translation parameters $(\beta_i,\theta_i,t_i)$, and $p_m$ the corresponding observed 3D
keypoint.
The purpose of this stage is to construct a case-specific and rig-consistent body frame that
anchors the organ representation. Full fitting
details, regularization terms, and SMPL-H-to-MHR conversion are provided in the  \hyperref[supp:methods]{Supplementary Methods}.

Internal organs are then canonicalized into the registered canonical rest-rig space to reduce pose-induced
variability before learning population priors. For each subject, the joint-wise
unposing transform
\begin{equation}
\label{eq:canonical-unpose-transform}
\Delta G_{i,j} = \left(G_{i,j}^{\mathrm{pose}}\right)^{-1} G_{i,j}^{\mathrm{rest}},
\end{equation}
where $G_{i,j}^{\mathrm{rest}}$ and $G_{i,j}^{\mathrm{pose}}$ denote the rest-rig and posed
joint transforms, respectively, under the row-vector affine convention defined in the
\hyperref[supp:canon_body]{Supplementary Methods},
is computed, and an approximate inverse-skinning operation is applied using distance-based blending weights
$w_{v,j}$:
\begin{equation}
\label{eq:inverse-skinning}
\tilde v^{\mathrm{canon}} = \sum_{j \in \mathcal{N}(v)} w_{v,j}\, \tilde v^{\mathrm{pose}}\, \Delta G_{i,j},
\qquad
\tilde v = [\,v\;\;1\,],
\qquad
\text{s.t. } \sum_{j \in \mathcal{N}(v)} w_{v,j}=1.
\end{equation}
where $v^{\mathrm{pose}}$ denotes a vertex in the posed/acquisition space,
$v^{\mathrm{canon}}$ its canonicalized counterpart in the subject-specific canonical rest-rig space, $\tilde
v^{\mathrm{pose}}$ and $\tilde v^{\mathrm{canon}}$ denote their augmented row-coordinate forms, and
$\mathcal{N}(v)$ the selected supporting joints for vertex $v$. This is an approximate
inverse-skinning (unposing) step: it applies a forward-style linear blend over joint-wise unposing
transforms rather than computing the exact inverse of the blended forward-skinning map.
To improve anatomical plausibility for internal organs, candidate joints are restricted to a
trunk-related subset, and each organ is supported by a small organ-centric joint set rather than
arbitrary whole-body joints. This organ-aware constraint reduces non-physical influence from distal
limbs and stabilizes the canonicalization of internal structures. After this step, organs are
expressed in a common canonical rest-rig space, but they still do not share topology across
subjects. Additional implementation details for weight assignment, trunk-related joint restriction,
scale handling, and geometric regularization are provided in the \hyperref[supp:methods]{Supplementary Methods}.

Because canonicalization resolves macroscopic pose discrepancies, dense correspondence in canonical space can be established by directly non-rigidly registering a semantic organ template to each canonicalized target organ. For organ class $k$, the deformed template is parameterized as
\begin{equation}
\label{eq:dense-corr-template-deformation}
\hat{V}_{i,k} = V_k + \Delta V_{i,k} + \mathbf{1}_{n_k}(t_{i,k}^{\mathrm{reg}})^\top,
\end{equation}
where $V_k$ is the canonical template, $\Delta V_{i,k}$ is a per-vertex offset field, and
$t_{i,k}^{\mathrm{reg}}$ is a global translation. The deformation is estimated under a compact
geometry-regularized objective,
\begin{equation}
\label{eq:dense-corr-objective}
E =
\lambda_{\mathrm{data}} E_{\mathrm{chamfer}}
+ \lambda_{\mathrm{edge}} E_{\mathrm{edge}}
+ \lambda_{\mathrm{normal}} E_{\mathrm{normal}}
+ \lambda_{\mathrm{lap}} E_{\mathrm{lap}}.
\end{equation}
These terms correspond to data fitting by Chamfer distance, edge-length regularization, normal
consistency, and Laplacian smoothness, respectively.
This step preserves template connectivity while adapting organ geometry to each subject, yielding a
set of canonical, body-aligned, and topology-consistent organ instances. These instances form the
representation basis for subsequent prior learning. Additional optimization details are provided in
the \hyperref[supp:methods]{Supplementary Methods}.

\subsubsection{Low-dimensional organ representation}\label{sec:methods-lowdim}
 
To reduce the coupling between anatomical individuality and pose-induced deformations, the
problem is formulated in a canonical rest-rig space. Following Section~\ref{sec:methods-canonical-construction}, a set
of canonical, topology-consistent organ instances alongside their corresponding rest-rig states 
is obtained for each organ class $k$. The training set is defined as
\begin{equation}
\label{eq:organ-training-set}
\mathcal{D}_k = \left\{ \left(T_{i,k}, G_i^{\mathrm{rest}}, \beta_i\right) \right\}_{i=1}^{N_k},
\end{equation}
where $T_{i,k}$ is the registered organ instance mesh of subject $i$ in canonical rest-rig space, 
$G_i^{\mathrm{rest}}$ is the corresponding rest-rig state, and $\beta_i$ denotes optional 
identity coefficients, with $N_k$ denoting the number of training instances for organ class $k$.
This representation provides a common basis for organ-wise prior learning across
subjects.

Importantly, the coordinate construction rule is shared globally across the body. Organ 
specificity arises from the organ template, the corresponding training instances, and the 
dynamically selected nearby skeletal features. An organ-wise mapping of the form
\begin{equation}
\label{eq:prior-mapping}
\left(\Delta c_{i,k},\, \ell_{i,k}\right) = \psi_k\left(G_i^{\mathrm{rest}}, \beta_i\right),
\end{equation}
is learned, where $\psi_k(\cdot)$ denotes the organ-wise mapping from subject-specific rest-rig
cues and optional identity coefficients to organ descriptors, $\Delta c_{i,k}$ is the local
centroid displacement of organ $k$, and $\ell_{i,k}$ its logarithmic anisotropic scale. The coarse
orientation prior $\bar R_k$ is estimated separately as an organ-wise mean orientation prior over
the training set rather than predicted as a subject-specific output. This formulation frames the prior learning as a 
spatial initialization problem: the model predicts how a template organ should be placed and 
scaled within the consistently defined frame, while providing uncertainty estimates for subsequent robotic 
refinement.

Given the canonical, topology-consistent organ instances obtained in
Section~\ref{sec:methods-canonical-construction}, the next objective is to obtain a representation
that is suitable for prior learning and for downstream Anatomical Representation Instantiation. The
goal is to encode each organ in a compact, template-preserving form that captures the aspects most
relevant to
robotic ultrasound initialization. In particular, the representation should retain where the organ
is, how far it extends, and its coarse geometric orientation in the shared body-aligned frame,
because these factors influence whether a valid acoustic window can be reached and whether a
proposed probe path is anatomically appropriate. This low-dimensional parameterization is also more
suitable for the present data regime, allowing organ-wise priors to be learned robustly without directly
regressing high-dimensional meshes.

To make organ placement comparable across subjects, all organ instances are expressed in a
subject-specific but consistently constructed ACS. For each subject,
a single body-centric ACS is derived from stable skeletal cues in the registered rest rig, so that
different organs are described under the same body-level spatial rule rather than under separate
organ-dependent frames. Let $(o_i^{\mathrm{ACS}}, R_i^{\mathrm{ACS}})$ denote the origin and
orientation of this anatomical frame for subject $i$. The canonical organ instance of class $k$ is
then mapped from world coordinates into the local ACS by
\begin{equation}
\label{eq:organ-to-local}
V_{i,k}^{\mathrm{local}} =
\bigl(V_{i,k}^{\mathrm{world}} - o_i^{\mathrm{ACS}}\bigr)
\bigl(R_i^{\mathrm{ACS}}\bigr)^\top.
\end{equation}
Using a shared ACS is important here because the subsequent prior is intended to model organ
placement and scale under a common anatomical reference, rather than to absorb arbitrary coordinate
differences across subjects.

The template is not redefined in each subject's local frame. Instead, it is fixed once in the
anatomical frame of a selected reference case,
\begin{equation}
\label{eq:reference-template-local}
V_{k,\mathrm{temp}}^{\mathrm{local}} = \bigl(V_{k,\mathrm{temp}}^{\mathrm{world}} -
o_{\mathrm{ref}}^{\mathrm{ACS}}\bigr) \bigl(R_{\mathrm{ref}}^{\mathrm{ACS}}\bigr)^\top,
\end{equation}
and this reference-local template is used throughout training. A rigid alignment is then computed
between the reference-local template and the subject-local organ using the Kabsch
algorithm~\cite{kabsch1976solution},
\begin{equation}
\label{eq:kabsch-alignment}
R_{i,k}, t_{i,k}^{\mathrm{rigid}} = \arg\min_{R \in SO(3),\, t}
\left\| V_{k,\mathrm{temp}}^{\mathrm{local}} R^\top + t - V_{i,k}^{\mathrm{local}} \right\|_F^2,
\end{equation}
which provides an instance-specific rotation descriptor together with an alignment error for
quality control. Importantly, although this alignment returns a rigid translation term
$t_{i,k}^{\mathrm{rigid}}$, the placement variable used for learning is not this rigid translation
itself. Instead, the centroid displacement in the shared anatomical frame is used,
\begin{equation}
\label{eq:centroid-displacement}
\Delta c_{i,k} = c\!\left(V_{i,k}^{\mathrm{local}}\right) -
c\!\left(V_{k,\mathrm{temp}}^{\mathrm{local}}\right),
\end{equation}
where $c(\cdot)$ denotes the vertex centroid. This choice is deliberate: after non-rigid template
registration, centroid displacement provides a more stable and anatomically intuitive description
of macroscopic organ placement than the rigid alignment translation alone, and is therefore better
matched to prior learning for initialization.

After rigid alignment, anisotropic scale is estimated by comparing the axis-wise spread of the 
aligned instance and the template. Specifically, the rigid alignment is first removed,
\begin{equation}
\label{eq:aligned-instance}
\tilde V_{i,k} = \bigl(V_{i,k}^{\mathrm{local}} - t_{i,k}^{\mathrm{rigid}}\bigr) R_{i,k},
\end{equation}
and the following quantity is then computed:
\begin{equation}
\label{eq:anisotropic-scale}
s_{i,k} =
\frac{
\mathrm{Std}_{\mathrm{axis}}\!\left(\tilde V_{i,k} - c(\tilde V_{i,k})\right)
}{
\mathrm{Std}_{\mathrm{axis}}\!\left(V_{k,\mathrm{temp}}^{\mathrm{local}} - c(V_{k,\mathrm{temp}}^{\mathrm{local}})\right)
+ \epsilon
},
\qquad
\ell_{i,k} = \log s_{i,k}.
\end{equation}
Here, $\epsilon > 0$ is a small constant for numerical stability.
In other words, each organ instance is reduced to template-relative location, scale, and
orientation descriptors in the shared ACS.
Thus, each organ instance is decomposed into the low-dimensional parameter set
\begin{equation}
\label{eq:lowdim-decomposition}
\{\Delta c_{i,k},\, R_{i,k},\, \ell_{i,k}\},
\end{equation}
where $\Delta c_{i,k}$ and $\ell_{i,k}$ are used as the learned targets, while $R_{i,k}$ is 
retained as an orientation descriptor.

To improve robustness, samples with abnormal alignment RMSE or volume ratio are removed by
interquartile-range filtering before regression.

This representation also defines the learning targets used in the next stage. Specifically,
$\Delta c_{i,k}$ and $\ell_{i,k}$ serve as the supervised targets for organ-wise prior learning,
whereas $R_{i,k}$ is retained as an orientation descriptor and summarized as an organ-wise mean
orientation prior. Consequently, the subsequent prior does not attempt to predict meshes directly.
Instead, it learns how local skeletal context explains template-relative organ placement and scale
within the shared anatomical frame.

\subsubsection{Skeleton-conditioned prior learning for organ placement and scale}\label{sec:methods-skeleton-prior}

\begin{figure*}[t]
\centering
\includegraphics[width=0.70\textwidth]{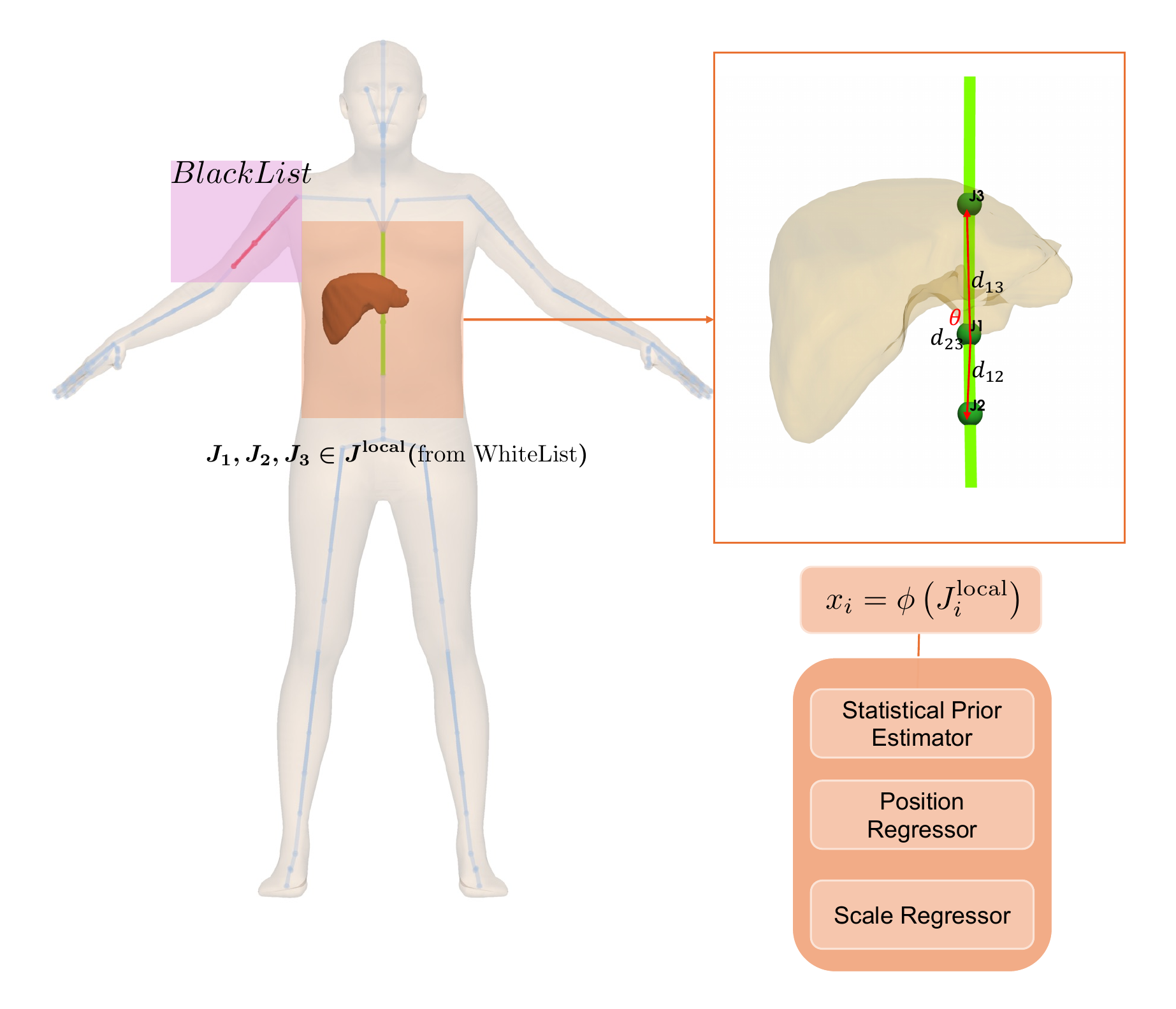}
\caption{Skeleton-conditioned prior regression. A local joint subset in rest space is converted to
explicit geometric features and fed to lightweight position and scale regressors.}
\label{fig443}
\end{figure*}

Given the low-dimensional organ representation defined in Section~\ref{sec:methods-lowdim}, prior learning is
formulated as predicting template-relative organ placement and scale from subject-specific skeletal
context. For each organ class $k$, the supervised targets are the centroid displacement
$\Delta c_{i,k}$ and the logarithmic anisotropic scale $\ell_{i,k}$, whereas the instance-specific
rotation descriptor $R_{i,k}$ is not learned as a subject-conditioned regression target. Instead,
these rotations are aggregated across the training set to define an organ-wise mean orientation
prior $\bar R_k$. This design keeps the learned prior focused on the quantities that are most
stable and most directly useful for initialization. The regression setup and feature flow are
summarized in Fig.~\ref{fig443}.

To predict these targets, we use explicit skeletal geometric features rather than learned latent
codes or dense body--organ shape spaces~\cite{guo2022smpl,shetty2023boss,kats2026depth,henrich2025looc}.
This choice follows naturally from the formulation in Section~\ref{sec:methods-lowdim}: because
the targets are explicit and low-dimensional, prior learning can be cast as regression over organ
placement and scale parameters rather than over mesh vertices or dense shape fields. The input
features are extracted from the subject-specific rest rig in the anatomical local frame. To preserve
anatomical relevance and maintain train--test consistency, each organ is associated with a small
fixed subset of joints selected from a predefined skeletal whitelist, and the same whitelist-defined
subset is used unchanged during both training and inference. Let
$J_{i,k}^{\mathrm{local}}$ denote the selected local joints for organ class $k$ in subject $i$.
An explicit feature vector is then defined as
\begin{equation}
\label{eq:skeletal-feature}
x_{i,k} = \phi\!\left(J_{i,k}^{\mathrm{local}}\right),
\end{equation}
where $\phi(\cdot)$ denotes the explicit skeletal feature extractor summarizing the local geometric
configuration of the selected joints. The whitelist construction rule, the organ-specific joint
definitions, and the detailed feature recipe are provided in the
\hyperref[supp:methods]{Supplementary Methods}.

For each organ class, lightweight regressors are fit for placement and scale,
\begin{equation}
\label{eq:placement-scale-regression}
\hat{\Delta c}_{i,k} = f_{\mathrm{pos},k}(x_{i,k}),
\qquad
\widehat{\ell}_{i,k} = f_{\mathrm{scale},k}(x_{i,k}).
\end{equation}
Here, $f_{\mathrm{pos},k}$ and $f_{\mathrm{scale},k}$ denote the organ-wise lightweight regressors
for placement and logarithmic scale, respectively.
The objective is to learn how local skeletal context explains organ placement and extent within the
shared anatomical frame, following the design
rationale discussed in Section~\ref{sec:methods-lowdim}.

After fitting, uncertainty is summarized at the level of the learned parameters rather than
as a full posterior over organ geometry. Specifically, residual variances are stored for centroid
displacement and logarithmic scale,
\begin{equation}
\label{eq:prior-residual-var}
\boldsymbol{\sigma}^2_{\mathrm{pos},k}
= \mathrm{Var}_{\mathrm{axis}}\!\left(\Delta c_{i,k} - \hat{\Delta c}_{i,k}\right),
\qquad
\boldsymbol{\sigma}^2_{\mathrm{scale},k}
= \mathrm{Var}_{\mathrm{axis}}\!\left(\ell_{i,k} - \widehat{\ell}_{i,k}\right),
\end{equation}
where $\mathrm{Var}_{\mathrm{axis}}(\cdot)$ denotes axis-wise residual variance,
together with the empirical covariance of centroid displacement,
\begin{equation}
\label{eq:placement-cov}
\Sigma_{\Delta c,k} = \mathrm{Cov}(\Delta c_{i,k}).
\end{equation}
where $\mathrm{Cov}(\cdot)$ denotes empirical covariance over training instances.
These quantities capture, respectively, conditional prediction uncertainty and population-level
placement variability in the anatomical frame.

The output of this stage is therefore a compact set of offline prior assets for each organ class:
placement and scale regressors, an organ-wise mean orientation prior, uncertainty statistics, and
the fixed whitelist-based feature definition required for train--test consistency. These assets are
used directly in the next stage to instantiate patient-specific anatomical hypotheses from external
body observations.

\subsection{Actionable Target Initialization: Control-facing geometric interface}\label{sec:methods-ati}

\begin{figure*}[t]
\centering
\includegraphics[width=0.70\textwidth]{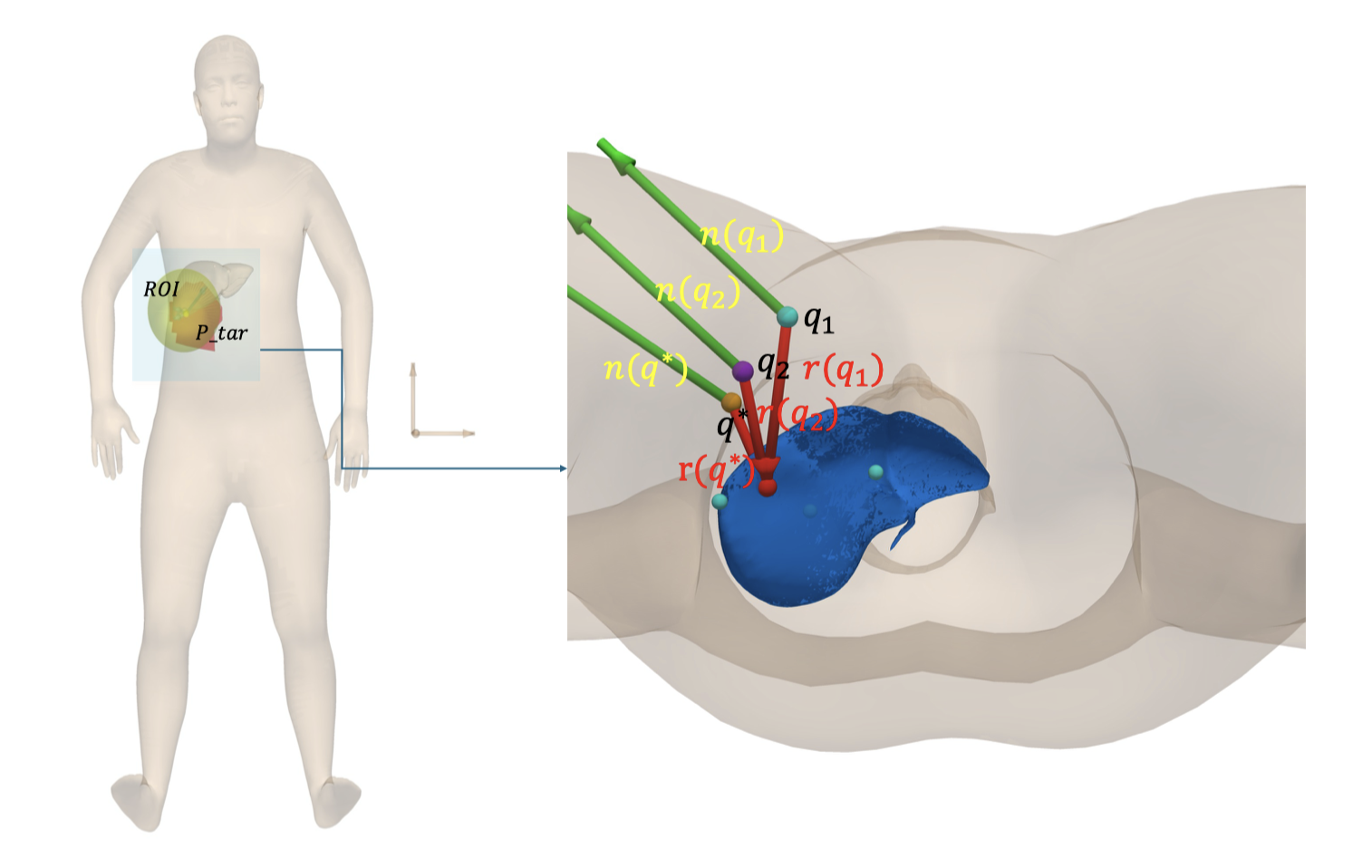}
\caption{Control-facing geometric interface. Anatomical targets are projected to candidate surface
contacts and optimized with normal-alignment, ray-alignment, and skeletal clearance constraints
for initialization-ready control signals.}
\label{fig47}
\end{figure*}

At inference time, the offline prior assets learned in
Section~\ref{sec:methods-skeleton-prior} are combined with the semantic target from
Section~\ref{sec:methods-semantic} and the current subject's body observations to instantiate a
subject-specific anatomical representation. In the current implementation, the position and scale
regressors are evaluated in the subject's anatomical coordinate system to obtain the instantiated
organ hypothesis. The control-facing geometric conversion summarized in Fig.~\ref{fig47} then maps
this instantiated anatomy to candidate contact states and initialization-ready probe geometry.
\begin{equation}
\label{eq:pred-local-organ}
\hat s_{i,k} = \exp(\widehat{\ell}_{i,k}),
\qquad
V_{i,k}^{\mathrm{pred,local}} = \Big((V_{k,\mathrm{temp}}^{\mathrm{local}} - \bar c_{k,\mathrm{temp}})
\odot \hat s_{i,k}\Big) \bar R_k^\top + \bar c_{k,\mathrm{temp}} + \hat{\Delta c}_{i,k},
\end{equation}
where $\hat s_{i,k}$ is the predicted anisotropic scale, $\odot$ denotes axis-wise scaling,
$\bar c_{k,\mathrm{temp}} = c\!\left(V_{k,\mathrm{temp}}^{\mathrm{local}}\right)$ is the
template centroid in the reference-local frame for organ class $k$, and $\bar R_k$ is the organ-wise mean orientation
prior estimated in Section~\ref{sec:methods-skeleton-prior} by aggregating the instance-specific rotation descriptors
$R_{i,k}$ over the training set; here $c(\cdot)$ denotes the centroid operator. In the present system, $\bar R_k$ acts as a conservative coarse
orientation cue rather than as a subject-specific predicted rotation. Let
$(o^{\mathrm{ACS}}, R^{\mathrm{ACS}})$ denote the origin and orientation of the current subject's
anatomical coordinate system. The instantiated organ hypothesis is then mapped to the subject's
world frame by
\begin{equation}
\label{eq:pred-world-organ}
V_{i,k}^{\mathrm{pred,world}} = V_{i,k}^{\mathrm{pred,local}} R^{\mathrm{ACS}} + o^{\mathrm{ACS}}.
\end{equation}
This procedure yields a deterministic anatomical representation that is geometrically consistent with the
learned template space, the subject-specific skeleton, and the semantic target specification.
For each initialization target, ATI resolves the grounded region specification $r$, together with
the inferred task type $\tau$, into an internal target point $p_{\mathrm{tar}}$ in the
instantiated anatomy.

Rather than returning anatomy as a passive reconstruction result, SAMe converts the instantiated
anatomical representation into control-facing geometric constraints through an anatomically constrained
contact optimization. Given an internal target point $p_{\mathrm{tar}} \in \mathbb{R}^3$---which
may correspond to an organ centroid, an ROI center, or an anatomical landmark---the objective is to
identify skin-surface contact candidates that maximize probe--target geometric alignment while
respecting anatomical feasibility. The procedure operates in two stages: seed initialization and
constrained candidate ranking.

For seed initialization, an initial surface point is obtained by forward projection along the
anterior body direction. Let $\mathbf{e}_z$ denote the anterior body direction in the current
anatomical frame. The projection ray is defined as
\begin{equation}
\label{eq:projection-ray}
\mathbf{r}_{\mathrm{proj}}(\lambda) = p_{\mathrm{tar}} + \lambda \, \mathbf{e}_z,
\qquad \lambda \ge 0,
\end{equation}
and its first intersection with the skin surface yields the seed point $p_{\mathrm{proj}}$. Around
$p_{\mathrm{proj}}$, a local forward-facing candidate set $\mathcal{Q}$ is constructed on the skin
surface.

Each candidate $q \in \mathcal{Q}$ is then evaluated by a composite score that combines geometric
alignment with anatomical feasibility. The alignment component measures how well the target-directed
unit ray opposes the outward surface normal:
\begin{equation}
\label{eq:ray-alignment}
\mathbf{r}(q) = \frac{p_{\mathrm{tar}} - q}{\left\| p_{\mathrm{tar}} - q \right\|_2},
\qquad
s_{\mathrm{align}}(q) = -\,\mathbf{n}(q)^\top \mathbf{r}(q).
\end{equation}
A second component penalizes candidates whose probe entry path intersects skeletal structures,
yielding a skeletal clearance term $s_{\mathrm{skel}}(q) \in [0, 1]$. The composite score is
\begin{equation}
\label{eq:composite-score}
s(q) = s_{\mathrm{align}}(q) \cdot s_{\mathrm{skel}}(q).
\end{equation}
Candidates with larger $s(q)$ simultaneously achieve better inward entry alignment and greater
skeletal clearance. The top $K_{\mathrm{cand}}$ candidates under this score are retained to form
$\mathcal{Q}_{\mathrm{top}}$, yielding candidate contact points together with their associated
surface normals, target-directed rays, and skeletal clearance metadata. The full vector
definitions, skeletal penalty formulation, and ranking details are provided in the
\hyperref[supp:methods]{Supplementary Methods}.

The abstract ATI output $z_p$ can therefore be instantiated as the following explicit structured
control-facing state,
\begin{equation}
\label{eq:control-state}
\zeta_k =
\left(
V_{i,k}^{\mathrm{pred,world}},
\mathcal{Q}_{\mathrm{top}},
\bar R_k,
\Sigma_{\Delta c,k},
\boldsymbol{\sigma}^2_{\mathrm{pos},k},
\boldsymbol{\sigma}^2_{\mathrm{scale},k}
\right),
\end{equation}
where $\mathcal{Q}_{\mathrm{top}}$ denotes the selected surface-contact candidates and the
remaining terms encode organ geometry and uncertainty in a shared frame. In the present
implementation, these candidates include their associated surface normals and target-directed rays.
Because this state is explicitly structured, future observation-conditioned refinement modules
could in principle operate on it as
\begin{equation}
\label{eq:future-refinement}
\zeta_k^{+} = \mathcal{F}(\zeta_k, y),
\end{equation}
where $y$ denotes additional observations acquired after initialization and $\mathcal{F}$ denotes a
possible future refinement operator. It is emphasized that no such online refinement mechanism is
instantiated or evaluated in the present study.

The skeletal clearance term $s_{\mathrm{skel}}(q)$ is computed from a subject-specific
bone-level skeleton prediction. Specifically, SKEL~\cite{keller2023skel}---a biomechanical
skeleton model originally designed for the SMPL-H body model---is adapted to the MHR rig through
a lightweight mapping from MHR parameters, enabling prediction of subject-specific skeletal
geometry including the rib cage and major bony landmarks. For each candidate $q$, the ray segment
from $q$ to $p_{\mathrm{tar}}$ is tested against the predicted skeletal surface. Let
$\mathcal{L}(q)$ denote this line segment, let $\mathcal{M}_{\mathrm{skel}}$ denote the predicted
skeletal surface, and let
\begin{equation}
\label{eq:skel-clearance}
d_{\mathrm{skel}}(q)=\min_{x \in \mathcal{L}(q),\, y \in \mathcal{M}_{\mathrm{skel}}}\|x-y\|_2,
\qquad
s_{\mathrm{skel}}(q)=\min\!\left(1,\frac{d_{\mathrm{skel}}(q)}{\delta_{\mathrm{skel}}}\right),
\end{equation}
where $\delta_{\mathrm{skel}} > 0$ is a fixed clearance threshold. Candidates whose entry paths
intersect bone therefore receive $s_{\mathrm{skel}}(q)=0$, whereas candidates with sufficient
clearance saturate at $s_{\mathrm{skel}}(q)=1$, steering the optimization away from acoustically
blocked access directions. This constraint is optional and can
be omitted when bone-level prediction is unavailable, in which case $s_{\mathrm{skel}}(q) = 1$ for
all candidates and the scoring reduces to geometric alignment alone.

\subsection{Inference-time system summary}\label{sec:methods-summary}

The preceding subsections describe the offline stages of SAMe, including semantic-prior
construction, canonical anatomical representation learning, and organ-wise prior estimation. At
inference time, these learned assets are combined with two upstream inputs: a clinical query or
task description $c$, and subject-specific body observations $x_b$ derived from external body
sensing. Clinical Semantics Grounding first maps $c$ to the grounded target tuple $(o,\tau,r)$ by
retrieving task-aligned semantic evidence and inferring the task type $\tau$. Anatomical
Representation Instantiation then applies the offline-learned placement and scale priors to $x_b$
to produce the patient-specific anatomical representation $z_a$. Finally, Actionable Target
Initialization uses $z_a$ together with body-surface cues $x_s \subseteq x_b$ to resolve the target
specification into internal target points and convert the instantiated anatomy into the abstract
initialization output $z_p$, realized in the present implementation as the control-facing
geometric state $\zeta_k$.

Downstream robotic modules therefore consume explicit geometric quantities rather than free-form
semantics or latent anatomy variables. In the present implementation, $\zeta_k$ comprises the
selected contact candidates $\mathcal{Q}_{\mathrm{top}}$, their associated normals and
target-directed rays, the instantiated organ geometry, and the accompanying uncertainty
descriptors, yielding an initialization-ready interface for contact-aware setup and local
image-driven execution. In compact form, the online chain is
$c \rightarrow (o,\tau,r) \rightarrow z_a \rightarrow z_p$, with $z_p$ realized in the present
implementation as $\zeta_k$.

\section*{Declarations}

\subsection*{Data availability}

The public and controlled-access datasets used in this study are described in Section~\ref{sec:methods-data}. The standard organ models and related weights used in this study are available with the code at the repository stated below. Additional real-robot ultrasound data are not publicly available owing to participant privacy and consent restrictions, but may be made available from the corresponding author (\href{mailto:jingzhang.cv@gmail.com}{jingzhang.cv@gmail.com}) upon reasonable request and subject to institutional approval.

\subsection*{Code availability}

The code for this project, together with the standard organ models and related weights used in this study, will be released at the GitHub repository: \url{https://github.com/MiliLab/Echo-SAMe}.




\bibliographystyle{naturemag}
\bibliography{ref}

@article{bi2024machine,
  title={Machine learning in robotic ultrasound imaging: Challenges and perspectives},
  author={Bi, Yuan and Jiang, Zhongliang and Duelmer, Felix and Huang, Dianye and Navab, Nassir},
  journal={Annual Review of Control, Robotics, and Autonomous Systems},
  volume={7},
  year={2024},
  publisher={Annual Reviews}
}

@article{liu2025screens,
  title={From screens to scenes: A survey of embodied AI in healthcare},
  author={Liu, Yihao and Cao, Xu and Chen, Tingting and Jiang, Yankai and You, Junjie and Wu, Minghua and Wang, Xiaosong and Feng, Mengling and Jin, Yaochu and Chen, Jintai},
  journal={Information Fusion},
  volume={119},
  pages={103033},
  year={2025},
  publisher={Elsevier}
}

@article{wang2024autonomous,
  title={Autonomous robotic system for carotid artery ultrasound scanning with visual servo navigation},
  author={Wang, Ziwen and Han, Yingying and Zhao, Baoliang and Xie, Haiqin and Yao, Liang and Li, Bing and Meng, Max Q-H and Hu, Ying},
  journal={IEEE Transactions on Medical Robotics and Bionics},
  volume={6},
  number={4},
  pages={1436--1447},
  year={2024},
  publisher={IEEE}
}

@article{jiang2023skeleton,
  title={Skeleton graph-based ultrasound-CT non-rigid registration},
  author={Jiang, Zhongliang and Li, Xuesong and Zhang, Chenyu and Bi, Yuan and Stechele, Walter and Navab, Nassir},
  journal={IEEE Robotics and Automation Letters},
  volume={8},
  number={8},
  pages={4394--4401},
  year={2023},
  publisher={IEEE}
}

@inproceedings{jiang2021motion,
  title={Motion-aware robotic 3D ultrasound},
  author={Jiang, Zhongliang and Wang, Hanyu and Li, Zhenyu and Grimm, Matthias and Zhou, Mingchuan and Eck, Ulrich and Brecht, Sandra V and Lueth, Tim C and Wendler, Thomas and Navab, Nassir},
  booktitle={2021 IEEE International Conference on Robotics and Automation (ICRA)},
  pages={12494--12500},
  year={2021},
  organization={IEEE},
  address={Xi'an, China}
}

@article{merouche2015robotic,
  title={A robotic ultrasound scanner for automatic vessel tracking and three-dimensional reconstruction of b-mode images},
  author={Merouche, Samir and Allard, Louise and Montagnon, Emmanuel and Soulez, Gilles and Bigras, Pascal and Cloutier, Guy},
  journal={IEEE transactions on ultrasonics, ferroelectrics, and frequency control},
  volume={63},
  number={1},
  pages={35--46},
  year={2015},
  publisher={IEEE}
}

@article{jiang2021autonomous,
  title={Autonomous robotic screening of tubular structures based only on real-time ultrasound imaging feedback},
  author={Jiang, Zhongliang and Li, Zhenyu and Grimm, Matthias and Zhou, Mingchuan and Esposito, Marco and Wein, Wolfgang and Stechele, Walter and Wendler, Thomas and Navab, Nassir},
  journal={IEEE Transactions on Industrial Electronics},
  volume={69},
  number={7},
  pages={7064--7075},
  year={2021},
  publisher={IEEE}
}

@inproceedings{chen2023fully,
  title={Fully robotized 3d ultrasound image acquisition for artery},
  author={Chen, Mingcong and Huang, Yuanrui and Chen, Jian and Zhou, Tongxi and Chen, Jiuan and Liu, Hongbin},
  booktitle={2023 IEEE International Conference on Robotics and Automation (ICRA)},
  pages={2690--2696},
  year={2023},
  organization={IEEE},
  address={London, UK}
}

@inproceedings{akbari2021robot,
  title={Robot-assisted breast ultrasound scanning using geometrical analysis of the seroma and image segmentation},
  author={Akbari, Mojtaba and Carriere, Jay and Sloboda, Ron and Meyer, Tyler and Usmani, Nawaid and Husain, Siraj and Tavakoli, Mahdi},
  booktitle={2021 IEEE/RSJ International Conference on Intelligent Robots and Systems (IROS)},
  pages={3784--3791},
  year={2021},
  organization={IEEE},
  address={Prague, Czech Republic}
}

@article{shetty2023boss,
  author    = "Shetty, K. and Birkhold, A. and Jaganathan, S. and Strobel, N. and Egger, B. and Kowarschik, M. and Maier, A.",
  title     = "{BOSS}: Bones, organs and skin shape model",
  journal   = "Computers in Biology and Medicine",
  volume    = "165",
  pages     = "107383",
  year      = "2023",
  publisher = "Elsevier",
  doi       = "10.1016/j.compbiomed.2023.107383"
}

@misc{selfridge2023healthy,
  doi       = {10.7937/NC7Z-4F76},
  url       = {https://www.cancerimagingarchive.net/collection/healthy-total-body-cts/},
  author    = {Selfridge, Aaron R. and Spencer, Benjamin and Shiyam Sundar, Lalith Kumar and Abdelhafez, Yasser and Nardo, Lorenzo and Cherry, Simon R. and Badawi, Ramsey D.},
  title     = {Low-Dose CT Images of Healthy Cohort (Healthy-Total-Body-CTs)},
  publisher = {The Cancer Imaging Archive},
  year      = {2023},
  copyright = {NIH Controlled Data Access Policy}
}

@incollection{lorensen1998marching,
  title={Marching cubes: A high resolution 3D surface construction algorithm},
  author={Lorensen, William E and Cline, Harvey E},
  booktitle={Seminal graphics: pioneering efforts that shaped the field},
  editor={Kaufman, Arie},
  pages={347--353},
  year={1998},
  publisher={ACM Press},
  address={New York, NY, USA}
}

@article{kabsch1976solution,
  title={A solution for the best rotation to relate two sets of vectors},
  author={Kabsch, Wolfgang},
  journal={Acta Crystallographica Section A},
  volume={32},
  number={5},
  pages={922--923},
  year={1976},
  publisher={International Union of Crystallography}
}

@inproceedings{hase2020ultrasound,
  title={Ultrasound-guided robotic navigation with deep reinforcement learning},
  author={Hase, Hannes and Azampour, Mohammad Farid and Tirindelli, Maria and Paschali, Magdalini and Simson, Walter and Fatemizadeh, Emad and Navab, Nassir},
  booktitle={2020 IEEE/RSJ International Conference on Intelligent Robots and Systems (IROS)},
  pages={5534--5541},
  year={2020},
  organization={IEEE}
}

@inproceedings{li2021autonomous,
  title={Autonomous navigation of an ultrasound probe towards standard scan planes with deep reinforcement learning},
  author={Li, Keyu and Wang, Jian and Xu, Yangxin and Qin, Hao and Liu, Dongsheng and Liu, Li and Meng, Max Q-H},
  booktitle={2021 IEEE International Conference on Robotics and Automation (ICRA)},
  pages={8302--8308},
  year={2021},
  organization={IEEE}
}

@article{ning2021autonomic,
  title={Autonomic robotic ultrasound imaging system based on reinforcement learning},
  author={Ning, Guochen and Zhang, Xinran and Liao, Hongen},
  journal={IEEE transactions on biomedical engineering},
  volume={68},
  number={9},
  pages={2787--2797},
  year={2021},
  publisher={IEEE}
}

@article{bi2022vesnet,
  title={VesNet-RL: Simulation-based reinforcement learning for real-world US probe navigation},
  author={Bi, Yuan and Jiang, Zhongliang and Gao, Yuan and Wendler, Thomas and Karlas, Angelos and Navab, Nassir},
  journal={IEEE Robotics and Automation Letters},
  volume={7},
  number={3},
  pages={6638--6645},
  year={2022},
  publisher={IEEE}
}

@inproceedings{droste2020automatic,
  title={Automatic probe movement guidance for freehand obstetric ultrasound},
  author={Droste, Richard and Drukker, Lior and Papageorghiou, Aris T and Noble, J Alison},
  booktitle={International Conference on Medical Image Computing and Computer-Assisted Intervention},
  pages={583--592},
  year={2020},
  organization={Springer}
}

@inproceedings{men2022multimodal,
  title={Multimodal-GuideNet: Gaze-probe bidirectional guidance in obstetric ultrasound scanning},
  author={Men, Qianhui and Teng, Clare and Drukker, Lior and Papageorghiou, Aris T and Noble, J Alison},
  booktitle={International Conference on Medical Image Computing and Computer-Assisted Intervention},
  pages={94--103},
  year={2022},
  organization={Springer}
}

@inproceedings{deng2021learning,
  title={Learning robotic ultrasound scanning skills via human demonstrations and guided explorations},
  author={Deng, Xutian and Chen, Yiting and Chen, Fei and Li, Miao},
  booktitle={2021 IEEE International Conference on Robotics and Biomimetics (ROBIO)},
  pages={372--378},
  year={2021},
  organization={IEEE}
}

@article{jiang2022towards,
  title={Towards autonomous atlas-based ultrasound acquisitions in presence of articulated motion},
  author={Jiang, Zhongliang and Gao, Yuan and Xie, Le and Navab, Nassir},
  journal={IEEE Robotics and Automation Letters},
  volume={7},
  number={3},
  pages={7423--7430},
  year={2022},
  publisher={IEEE}
}

@article{hennersperger2016towards,
  title={Towards MRI-based autonomous robotic US acquisitions: a first feasibility study},
  author={Hennersperger, Christoph and Fuerst, Bernhard and Virga, Salvatore and Zettinig, Oliver and Frisch, Benjamin and Neff, Thomas and Navab, Nassir},
  journal={IEEE transactions on medical imaging},
  volume={36},
  number={2},
  pages={538--548},
  year={2016},
  publisher={IEEE}
}

@article{jiang2022precise,
  title={Precise repositioning of robotic ultrasound: Improving registration-based motion compensation using ultrasound confidence optimization},
  author={Jiang, Zhongliang and Danis, Nehil and Bi, Yuan and Zhou, Mingchuan and Kroenke, Markus and Wendler, Thomas and Navab, Nassir},
  journal={IEEE Transactions on Instrumentation and Measurement},
  volume={71},
  pages={1--11},
  year={2022},
  publisher={IEEE}
}

@article{jiang2025towards,
  title={Towards expert-level autonomous carotid ultrasonography with large-scale learning-based robotic system},
  author={Jiang, Haojun and Zhao, Andrew and Yang, Qian and Yan, Xiangjie and Wang, Teng and Wang, Yulin and Jia, Ning and Wang, Jiangshan and Wu, Guokun and Yue, Yang and others},
  journal={Nature Communications},
  volume={16},
  number={1},
  pages={7893},
  year={2025},
  publisher={Nature Publishing Group UK London}
}

@article{PhysioNet-mimiciv-2.2,
  author = {Johnson, Alistair and Bulgarelli, Lucas and Pollard, Tom and Horng, Steven and Celi, Leo Anthony and Mark, Roger},
  title = {{MIMIC-IV}},
  journal = {{PhysioNet}},
  year = {2023},
  month = jan,
  note = {Version 2.2},
  doi = {10.13026/6mm1-ek67},
  url = {https://doi.org/10.13026/6mm1-ek67}
}

@article{gutschmayer2025whole,
  title={Whole-Body [18F] FDG-PET/CT Imaging of Healthy Controls: Test/Retest Data for Systemic, Multi-Organ Analysis},
  author={Gutschmayer, Sebastian and Yu, Josef and Geist, Barbara Katharina and {\"O}zer, {\"O}yk{\"u} and Reiterits, Bettina and Ferrara, Daria and Pires, Manuel and Rausch, Ivo and Ibeschitz, Harald and Karanikas, Georgios and others},
  journal={Scientific Data},
  volume={12},
  number={1},
  pages={1707},
  year={2025},
  publisher={Nature Publishing Group UK London},
  doi={10.1038/s41597-025-05997-4}
}

@misc{quadrahc_zenodo2025,
  author={Gutschmayer, Sebastian and Yu, Josef and others},
  title={Whole-Body [18F]FDG-PET/CT Imaging of Healthy Controls: Test/Retest Data for Systemic, Multi-Organ Analysis},
  year={2025},
  publisher={Zenodo},
  doi={10.5281/zenodo.16364694},
  url={https://doi.org/10.5281/zenodo.16364694}
}

@article{ferguson2025mhr,
  title={Mhr: Momentum human rig},
  author={Ferguson, Aaron and Osman, Ahmed AA and Bescos, Berta and Stoll, Carsten and Twigg, Chris and Lassner, Christoph and Otte, David and Vignola, Eric and Prada, Fabian and Bogo, Federica and others},
  journal={arXiv preprint arXiv:2511.15586},
  year={2025}
}

@article{yang2026sam,
  title={Sam 3d body: Robust full-body human mesh recovery},
  author={Yang, Xitong and Kukreja, Devansh and Pinkus, Don and Sagar, Anushka and Fan, Taosha and Park, Jinhyung and Shin, Soyong and Cao, Jinkun and Liu, Jiawei and Ugrinovic, Nicolas and others},
  journal={arXiv preprint arXiv:2602.15989},
  year={2026}
}

@misc{MHR:2025,
  title={MHR: Momentum Human Rig},
  author={Aaron Ferguson and Ahmed A. A. Osman and Berta Bescos and Carsten Stoll and Chris Twigg and Christoph Lassner and David Otte and Eric Vignola and Fabian Prada and Federica Bogo and Igor Santesteban and Javier Romero and Jenna Zarate and Jeongseok Lee and Jinhyung Park and Jinlong Yang and John Doublestein and Kishore Venkateshan and Kris Kitani and Ladislav Kavan and Marco Dal Farra and Matthew Hu and Matthew Cioffi and Michael Fabris and Michael Ranieri and Mohammad Modarres and Petr Kadlecek and Rawal Khirodkar and Rinat Abdrashitov and Romain Pr{\'e}vost and Roman Rajbhandari and Ronald Mallet and Russell Pearsall and Sandy Kao and Sanjeev Kumar and Scott Parrish and Shoou-I Yu and Shunsuke Saito and Takaaki Shiratori and Te-Li Wang and Tony Tung and Yichen Xu and Yuan Dong and Yuhua Chen and Yuanlu Xu and Yuting Ye and Zhongshi Jiang},
  year={2025},
  eprint={2511.15586},
  archivePrefix={arXiv},
  primaryClass={cs.GR},
  url={https://arxiv.org/abs/2511.15586}
}

@article{henrich2025looc,
  title={Looc: Localizing organs using occupancy networks and body surface depth images},
  author={Henrich, Pit and Mathis-Ullrich, Franziska},
  journal={IEEE Access},
  year={2025},
  publisher={IEEE}
}

@inproceedings{guo2022smpl,
  title={SMPL-A: Modeling person-specific deformable anatomy},
  author={Guo, Hengtao and Planche, Benjamin and Zheng, Meng and Karanam, Srikrishna and Chen, Terrence and Wu, Ziyan},
  booktitle={Proceedings of the IEEE/CVF Conference on Computer Vision and Pattern Recognition},
  pages={20814--20823},
  year={2022}
}

@article{kats2026depth,
  title={Depth to Anatomy: Learning Internal Organ Locations from Surface Depth Images},
  author={Kats, Eytan and Geissler, Kai and Mensing, Daniel and Hirsch, Jochen G and Heldman, Stefan and Heinrich, Mattias P},
  journal={arXiv preprint arXiv:2601.18260},
  year={2026}
}

@article{su2024fully,
  title={A fully autonomous robotic ultrasound system for thyroid scanning},
  author={Su, Kang and Liu, Jingwei and Ren, Xiaoqi and Huo, Yingxiang and Du, Guanglong and Zhao, Wei and Wang, Xueqian and Liang, Bin and Li, Di and Liu, Peter Xiaoping},
  journal={Nature communications},
  volume={15},
  number={1},
  pages={4004},
  year={2024},
  publisher={Nature Publishing Group UK London}
}

@inproceedings{long2024localizing,
  title={Localizing Scan Targets from Human Pose for Autonomous Lung Ultrasound Imaging},
  author={Long, Jianzhi and Cai, Jicang and Al-Battal, Abdullah and Jin, Shiwei and Zhang, Jing and Tao, Dacheng and Lerman, Imanuel and Nguyen, Truong},
  booktitle={Intelligent Systems Conference},
  pages={610--625},
  year={2024},
  organization={Springer}
}

@article{xu2022vitpose,
  title={Vitpose: Simple vision transformer baselines for human pose estimation},
  author={Xu, Yufei and Zhang, Jing and Zhang, Qiming and Tao, Dacheng},
  journal={Advances in neural information processing systems},
  volume={35},
  pages={38571--38584},
  year={2022}
}

@article{johnson2019billion,
  title={Billion-scale similarity search with {GPUs}},
  author={Johnson, Jeff and Douze, Matthijs and J{\'e}gou, Herv{\'e}},
  journal={IEEE Transactions on Big Data},
  volume={7},
  number={3},
  pages={535--547},
  year={2019},
  publisher={IEEE}
}

@inproceedings{SMPL-X:2019,
    title = {Expressive Body Capture: 3D Hands, Face, and Body from a Single Image},
    author = {Pavlakos, Georgios and Choutas, Vasileios and Ghorbani, Nima and Bolkart, Timo and Osman, Ahmed A. A. and Tzionas, Dimitrios and Black, Michael J.},
    booktitle = {Proceedings IEEE Conf. on Computer Vision and Pattern Recognition (CVPR)},
    year = {2019}
}

@article{MANO:SIGGRAPHASIA:2017,
    title = {Embodied Hands: Modeling and Capturing Hands and Bodies Together},
    author = {Romero, Javier and Tzionas, Dimitrios and Black, Michael J.},
    journal = {ACM Transactions on Graphics, (Proc. SIGGRAPH Asia)},
    volume = {36},
    number = {6},
    series = {245:1--245:17},
    month = nov,
    year = {2017},
    month_numeric = {11}
  }

@article{SMPL:2015,
    author = {Loper, Matthew and Mahmood, Naureen and Romero, Javier and Pons-Moll, Gerard and Black, Michael J.},
    title = {{SMPL}: A Skinned Multi-Person Linear Model},
    journal = {ACM Transactions on Graphics, (Proc. SIGGRAPH Asia)},
    month = oct,
    number = {6},
    pages = {248:1--248:16},
    publisher = {ACM},
    volume = {34},
    year = {2015}
}

@article{salomon2011practice,
  title={Practice guidelines for performance of the routine mid-trimester fetal ultrasound scan.},
  author={Salomon, Laurent Julien and Alfirevic, Zarko and Berghella, Vincenzo and Bilardo, C and Hernandez-Andrade, Edgar and Johnsen, Synn{\o}ve Lian and Kalache, K and Leung, K-Y and Malinger, G and Munoz, H and others},
  journal={Ultrasound in Obstetrics \& Gynecology},
  volume={37},
  number={1},
  year={2011}
}

@article{namburete2023normative,
  title={Normative spatiotemporal fetal brain maturation with satisfactory development at 2 years},
  author={Namburete, Ana IL and Papie{\.z}, Bart{\l}omiej W and Fernandes, Michelle and Wyburd, Madeleine K and Hesse, Linde S and Moser, Felipe A and Ismail, Leila Cheikh and Gunier, Robert B and Squier, Waney and Ohuma, Eric O and others},
  journal={Nature},
  volume={623},
  number={7985},
  pages={106--114},
  year={2023},
  publisher={Nature Publishing Group UK London}
}

@article{ulloa2021deep,
  title={Deep-learning-assisted analysis of echocardiographic videos improves predictions of all-cause mortality},
  author={Ulloa Cerna, Alvaro E and Jing, Linyuan and Good, Christopher W and vanMaanen, David P and Raghunath, Sushravya and Suever, Jonathan D and Nevius, Christopher D and Wehner, Gregory J and Hartzel, Dustin N and Leader, Joseph B and others},
  journal={Nature Biomedical Engineering},
  volume={5},
  number={6},
  pages={546--554},
  year={2021},
  publisher={Nature Publishing Group UK London}
}

@article{stein2008use,
  title={Use of carotid ultrasound to identify subclinical vascular disease and evaluate cardiovascular disease risk: a consensus statement from the American Society of Echocardiography Carotid Intima-Media Thickness Task Force endorsed by the Society for Vascular Medicine},
  author={Stein, James H and Korcarz, Claudia E and Hurst, R Todd and Lonn, Eva and Kendall, Christopher B and Mohler, Emile R and Najjar, Samer S and Rembold, Christopher M and Post, Wendy S},
  journal={Journal of the American Society of echocardiography},
  volume={21},
  number={2},
  pages={93--111},
  year={2008},
  publisher={Elsevier}
}

@article{ferraioli2019ultrasound,
  title={Ultrasound-based techniques for the diagnosis of liver steatosis},
  author={Ferraioli, Giovanna and Monteiro, Livia Beatriz Soares},
  journal={World journal of gastroenterology},
  volume={25},
  number={40},
  pages={6053},
  year={2019}
}

@article{ferraioli2018liver,
  title={Liver ultrasound elastography: an update to the world federation for ultrasound in medicine and biology guidelines and recommendations},
  author={Ferraioli, Giovanna and Wong, Vincent Wai-Sun and Castera, Laurent and Berzigotti, Annalisa and Sporea, Ioan and Dietrich, Christoph F and Choi, Byung Ihn and Wilson, Stephanie R and Kudo, Masatoshi and Barr, Richard G},
  journal={Ultrasound in medicine \& biology},
  volume={44},
  number={12},
  pages={2419--2440},
  year={2018},
  publisher={Elsevier}
}

@article{tahmasebpour2005sonographic,
  title={Sonographic examination of the carotid arteries},
  author={Tahmasebpour, Hamid R and Buckley, Anne R and Cooperberg, Peter L and Fix, Cathy H},
  journal={Radiographics},
  volume={25},
  number={6},
  pages={1561--1575},
  year={2005},
  publisher={Radiological Society of North America}
}

@article{lin2024fully,
  title={A fully integrated wearable ultrasound system to monitor deep tissues in moving subjects},
  author={Lin, Muyang and Zhang, Ziyang and Gao, Xiaoxiang and Bian, Yizhou and Wu, Ray S and Park, Geonho and Lou, Zhiyuan and Zhang, Zhuorui and Xu, Xiangchen and Chen, Xiangjun and others},
  journal={Nature biotechnology},
  volume={42},
  number={3},
  pages={448--457},
  year={2024},
  publisher={Nature Publishing Group US New York}
}

@article{hu2023wearable,
  title={A wearable cardiac ultrasound imager},
  author={Hu, Hongjie and Huang, Hao and Li, Mohan and Gao, Xiaoxiang and Yin, Lu and Qi, Ruixiang and Wu, Ray S and Chen, Xiangjun and Ma, Yuxiang and Shi, Keren and others},
  journal={Nature},
  volume={613},
  number={7945},
  pages={667--675},
  year={2023},
  publisher={Nature Publishing Group UK London}
}

@article{wang2018monitoring,
  title={Monitoring of the central blood pressure waveform via a conformal ultrasonic device},
  author={Wang, Chonghe and Li, Xiaoshi and Hu, Hongjie and Zhang, Lin and Huang, Zhenlong and Lin, Muyang and Zhang, Zhuorui and Yin, Zhenan and Huang, Brady and Gong, Hua and others},
  journal={Nature biomedical engineering},
  volume={2},
  number={9},
  pages={687--695},
  year={2018},
  publisher={Nature Publishing Group UK London}
}

@article{won2024sound,
  title={Sound the alarm: the sonographer shortage is echoing across healthcare},
  author={Won, Daniel and Walker, James and Horowitz, Russ and Bharadwaj, Sandeep and Carlton, Edward and Gabriel, Helena},
  journal={Journal of ultrasound in medicine},
  volume={43},
  number={7},
  pages={1289--1301},
  year={2024},
  publisher={Wiley Online Library}
}

@techreport{asrt2023staffing_survey,  
author = {{American Society of Radiologic Technologists}},  
title = {2023 Radiologic Sciences Workplace and Staffing Survey},  
institution = {American Society of Radiologic Technologists},  
year = {2023},  
url = {https://www.asrt.org/docs/default-source/research/staffing-surveys/radiologic-sciences-workplace-and-staffing-survey-2023.pdf},  
urldate = {2026-04-14}  
}

@misc{bls2025diagnostic_sonographers,  
author = {{Bureau of Labor Statistics, U.S. Department of Labor}},  
title = {Diagnostic Medical Sonographers},  
howpublished = {Occupational Outlook Handbook},  
year = {2025},  
month = aug,  
note = {Last modified August 28, 2025},  
url = {https://www.bls.gov/ooh/healthcare/diagnostic-medical-sonographers.htm},  
urldate = {2026-04-14}  
}

@article{mcgregor2020providing,
  title={Providing a sustainable sonographer workforce in Australia: clinical training solutions},
  author={McGregor, Rod and Pollard, Karen and Davidson, Rob and Moss, Cameron},
  journal={Sonography},
  volume={7},
  number={4},
  pages={141--147},
  year={2020},
  publisher={Wiley Online Library}
}

@techreport{sonographycanada2023hhr_submission,  
author = {{Sonography Canada}},  
title = {Addressing Health Human Resource Issues in Diagnostic Medical Imaging -- Sonography: Pre-Budget Submission for the 2024--25 Federal Budget},  
institution = {Sonography Canada},  
year = {2023},  
month = oct,  
url = {https://sonographycanada.ca/app/uploads/2023/10/Sonography-Canada-Pre-Budget-Final.pdf},  
urldate = {2026-04-14}  
}

@misc{sor2026_ultrasound_vacancy,
  author       = {{Society of Radiographers}},
  title        = {Sonography vacancy rates have increased dramatically, SoR ultrasound census reveals},
  year         = {2026},
  month        = mar,
  day          = {30},
  url          = {https://www.sor.org/news/ultrasound/sonography-vacancy-rates-have-increased-dramatical},
  note         = {Accessed 14 Apr 2026}
}

@article{parker2015educating,
  title={Educating the future sonographic workforce: membership survey report from the British Medical Ultrasound Society},
  author={Parker, PC and Harrison, G},
  journal={Ultrasound},
  volume={23},
  number={4},
  pages={231--241},
  year={2015},
  publisher={SAGE Publications Sage UK: London, England}
}

@article{beales2011reproducibility,
  title={Reproducibility of ultrasound measurement of the abdominal aorta},
  author={Beales, L and Wolstenhulme, Stephen and Evans, JA and West, Robert and Scott, DJA},
  journal={Journal of British Surgery},
  volume={98},
  number={11},
  pages={1517--1525},
  year={2011},
  publisher={Oxford University Press}
}

@article{joakimsen1997reproducibility,
  title={Reproducibility of ultrasound assessment of carotid plaque occurrence, thickness, and morphology: the Troms{\o} Study},
  author={Joakimsen, Oddmund and B{\o}naa, Kaare H and Stensland-Bugge, Eva},
  journal={Stroke},
  volume={28},
  number={11},
  pages={2201--2207},
  year={1997},
  publisher={Lippincott Williams \& Wilkins}
}

@article{kojcev2017reproducibility,
  title={On the reproducibility of expert-operated and robotic ultrasound acquisitions},
  author={Kojcev, Risto and Khakzar, Ashkan and Fuerst, Bernhard and Zettinig, Oliver and Fahkry, Carole and DeJong, Robert and Richmon, Jeremy and Taylor, Russell and Sinibaldi, Edoardo and Navab, Nassir},
  journal={International journal of computer assisted radiology and surgery},
  volume={12},
  number={6},
  pages={1003--1011},
  year={2017},
  publisher={Springer}
}

@article{coleman2024exploring,
  title={Exploring UK sonographers’ views on the use of professional supervision in clinical practice--Stage one findings of a mixed method study},
  author={Coleman, Gillian and Hyde, Emma and Strudwick, Ruth},
  journal={Radiography},
  volume={30},
  number={1},
  pages={252--256},
  year={2024},
  publisher={Elsevier}
}

@article{swerdlow2017robotic,
  title={Robotic arm--assisted sonography: Review of technical developments and potential clinical applications},
  author={Swerdlow, Daniel R and Cleary, Kevin and Wilson, Emmanuel and Azizi-Koutenaei, Bamshad and Monfaredi, Reza},
  journal={American Journal of Roentgenology},
  volume={208},
  number={4},
  pages={733--738},
  year={2017},
  publisher={American Roentgen Ray Society}
}

@article{monfaredi2015robot,
  title={Robot-assisted ultrasound imaging: Overview and development of a parallel telerobotic system},
  author={Monfaredi, Reza and Wilson, Emmanuel and Azizi koutenaei, Bamshad and Labrecque, Brendan and Leroy, Kristen and Goldie, James and Louis, Eric and Swerdlow, Daniel and Cleary, Kevin},
  journal={Minimally Invasive Therapy \& Allied Technologies},
  volume={24},
  number={1},
  pages={54--62},
  year={2015},
  publisher={Taylor \& Francis}
}

@article{leenhardt20132013,
  title={2013 European thyroid association guidelines for cervical ultrasound scan and ultrasound-guided techniques in the postoperative management of patients with thyroid cancer},
  author={Leenhardt, L and Erdogan, MF and Hegedus, L and Mandel, SJ and Paschke, R and Rago, T and Russ, G},
  journal={European thyroid journal},
  volume={2},
  number={3},
  pages={147--159},
  year={2013},
  publisher={S. Karger AG}
}

@article{huang2023review,
  title={Review of robot-assisted medical ultrasound imaging systems: Technology and clinical applications},
  author={Huang, Qinghua and Zhou, Jiakang and Li, ZhiJun},
  journal={Neurocomputing},
  volume={559},
  pages={126790},
  year={2023},
  publisher={Elsevier}
}

@article{du2024review,
  title={A review of robot-assisted ultrasound examination: Systems and technology},
  author={Du, Haiyan and Zhang, Xinran and Zhang, Yongde and Zhang, Fujun and Lin, Letao and Huang, Tao},
  journal={The International Journal of Medical Robotics and Computer Assisted Surgery},
  volume={20},
  number={4},
  pages={e2660},
  year={2024},
  publisher={Wiley Online Library}
}

@inproceedings{keller2023skel,
  title = {From Skin to Skeleton: Towards Biomechanically Accurate 3D Digital Humans},
  author = {Keller, Marilyn and Werling, Keenon and Shin, Soyong and Delp, Scott and Pujades, Sergi and Liu, C. Karen and Black, Michael J.},
  booktitle = {ACM ToG, Proc.~SIGGRAPH Asia},
  volume = {42},
  number = {6},
  month = dec,
  year = {2023},
}

@book{willcox2023foundational,
  title={Foundational research gaps and future directions for digital twins},
  author={Willcox, Karen and Bingham, D and Chung, C and Chung, J and Cruz-Neira, C and Grant, C and Kinter, J and Leung, R and Moin, P and Ohno-Machado, L and others},
  year={2023},
  publisher={National Academies Press Washington, DC, USA}
}

@article{tudor2025scoping,
  title={A scoping review of human digital twins in healthcare applications and usage patterns},
  author={Tudor, Brant H and Shargo, Ryan and Gray, Geoffrey M and Fierstein, Jamie L and Kuo, Frederick H and Burton, Robert and Johnson, Joyce T and Scully, Brandi B and Asante-Korang, Alfred and Rehman, Mohamed A and others},
  journal={npj Digital Medicine},
  volume={8},
  number={1},
  pages={587},
  year={2025},
  publisher={Nature Publishing Group UK London}
}

@article{drummond2024definitions,
  title={Definitions and characteristics of patient digital twins being developed for clinical use: scoping review},
  author={Drummond, David and Gonsard, Apolline},
  journal={Journal of Medical Internet Research},
  volume={26},
  pages={e58504},
  year={2024},
  publisher={JMIR Publications Toronto, Canada}
}

@article{koratala2019point,
  title={Point of care renal ultrasonography for the busy nephrologist: a pictorial review},
  author={Koratala, Abhilash and Bhattacharya, Deepti and Kazory, Amir},
  journal={World Journal of Nephrology},
  volume={8},
  number={3},
  pages={44},
  year={2019}
}

@article{scalea1999focused,
  title={Focused assessment with sonography for trauma (FAST): results from an international consensus conference},
  author={Scalea, Thomas M and Rodriguez, Aurelio and Chiu, William C and Brenneman, Frederick D and Fallon, Willaim F and Kato, Kazuyoshi and McKenney, Mark G and Nerlich, Michael L and Ochsner, M Gage and Yoshii, Hiroshi},
  journal={Journal of Trauma and Acute Care Surgery},
  volume={46},
  number={3},
  pages={466--472},
  year={1999},
  publisher={LWW}
}

@article{rozycki1998early,
  title={Early detection of hemoperitoneum by ultrasound examination of the right upper quadrant: a multicenter study},
  author={Rozycki, Grace S and Ochsner, M Gage and Feliciano, David V and Thomas, Bruce and Boulanger, Bernard R and Davis, Frank E and Falcone, Robert E and Schmidt, Judith A},
  journal={Journal of Trauma and Acute Care Surgery},
  volume={45},
  number={5},
  pages={878--883},
  year={1998},
  publisher={LWW}
}

\section*{Acknowledgements}

This work was supported in part by the New Generation Artificial Intelligence-National Science and Technology Major Project under Grant No.~2025ZD0123602.

\section*{Author contributions}

J.Z. and D.C. conceived the study, developed the methodology, and designed the experiments. J.Z. contributed to project administration, supervision and funding acquisition. D.C. was the primary developer of SAMe and led the software implementation, data curation and processing, computational and real-robot experiments, quantitative analysis, and figure preparation. W.J. contributed to results analysis and manuscript revision. Z.L. assisted with experimental execution, data collection, and investigation. J.L. contributed clinical guidance, medical advice, and expert evaluation of the ultrasound experimental results. Q.Z. and X.C. contributed clinical guidance and medical advice. B.D. contributed to conceptualization and supervision. C.F.D. contributed to clinical conceptualization, clinical and ultrasound-medicine guidance. D.T. contributed to conceptualization, provided expert guidance on artificial intelligence methodology and system-level framing, and contributed to senior supervision. D.C. and J.Z. wrote the initial manuscript. All authors reviewed, edited and approved the final manuscript. J.Z. and D.C. contributed equally to this work.

\section*{Competing interests}

The authors declare no competing interests.

\clearpage  
\section*{Supplementary information}
\addcontentsline{toc}{section}{Supplementary information}  

\section*{Supplementary Methods}\label{supp:methods}
\setcounter{table}{0}
\renewcommand{\thetable}{S\arabic{table}}
\renewcommand{\theHtable}{supp.\arabic{table}}

\subsection*{CT surface reconstruction and preprocessing}\label{supp:ct_recon}

For the CT data described in Section~\ref{sec:methods-data}, we reconstructed surfaces from the available coarse multi-tissue labels and designated 11 ultrasound-relevant anatomical structures as the core asset
set for subsequent atlas construction and localization experiments: aorta, bladder, heart, left
kidney, right kidney, liver, lung, pancreas, spleen, thyroid, and inferior vena cava. Each 3D
annotated volume was partitioned into four anatomically motivated surface sets: a skin envelope
$S_i$, a skeletal set $B_i$, a core asset set $O^c_i$, and an auxiliary organ set $O^a_i$,
where $O^c_i$ contains the ultrasound-relevant anatomical structures used in subsequent non-rigid
atlas construction and $O^a_i$ aggregates the remaining annotated anatomy.

Following prior anatomical shape modeling pipelines \cite{shetty2023boss}, segmentations were
converted to surface meshes using Marching Cubes \cite{lorensen1998marching} independently for
each anatomical label, with a sampling step size of 2 (and \texttt{skin\_step\_size}=2 for the
skin surface). Prior to isosurface extraction, light volumetric smoothing was applied (Gaussian
$\sigma=1.2$ for skin and $\sigma=1.0$ for organs); isosurfaces were extracted at levels 0.45
(skin) and 0.5 (organs). Minimal validity checks were applied to facilitate downstream registration
and simulation readiness, including connected-component filtering, screening for non-manifold
elements and self-intersections, and label-size filtering (\texttt{min\_voxels}=5000 for organs;
\texttt{skin\_min\_voxels}=10000).

\subsection*{Mesh assets for canonical anatomical representation}\label{supp:mesh_assets}

As described in Section~\ref{sec:methods-data}, the reconstructed CT surfaces serve as the geometric input to the
canonical anatomical representation pipeline in Section~\ref{sec:methods-canonical}. For each subject $i$, the
case-specific mesh assets are denoted as
\begin{equation}
\mathcal{A}_i = \left\{ M_i^{\mathrm{skin}}, M_{i,1}^{\mathrm{organ}}, M_{i,2}^{\mathrm{organ}}, \ldots \right\},
\qquad M = (V, F),
\end{equation}
where $V \subset \mathbb{R}^3$ and $F$ are the vertex and triangular face sets. These meshes are
surface-based and case-specific, preserve the posed anatomy at acquisition, and are grounded in the
CT world coordinate system via the NIfTI affine transform. At this stage, they are not topology-consistent
across subjects: meshes may differ in resolution and vertex indexing, and no dense correspondence
is assumed. The asset categories used in this stage are summarized in
Table~\ref{tab:mesh_assets}.

\begin{table}[t]
\centering
\caption{Summary of case-specific mesh assets used for canonical anatomical representation.}
\label{tab:mesh_assets}%
\begin{tabular}{@{}lp{0.68\columnwidth}@{}}
\toprule
Item & Description \\
\midrule
Data source & Healthy whole-body CT segmentations (TCIA) \\
Reconstruction & Label volumes $\rightarrow$ surface meshes via Marching Cubes \\
Assets & Skin mesh and multiple organ meshes per case \\
Semantics & Explicit anatomical labels for skin and organs \\
Coordinate frame & CT world coordinate frame (NIfTI affine) \\
Pose state & Posed anatomy at acquisition \\
Topology & No shared topology or dense correspondence across cases \\
\bottomrule
\end{tabular}
\end{table}

\subsection*{Semantic layer: structured unit definition, staged distillation, and retrieval records}\label{supp:semantic_layer}

Building on the final-study clinical text corpus formally defined in Section~\ref{sec:methods-data-text}, a semantic prior
database was constructed for task grounding. Rather than using the corpus as raw free-text evidence alone, it
was distilled into structured symptom--diagnosis--organ--anatomical-location units that can be
indexed and retrieved as anatomy-aware semantic priors.

Each distilled record was represented as a structured semantic unit
$u=\left(s,d,o,a,b\right)$, consisting of a dominant symptom/sign $s$, a diagnosis anchor $d$, a
standardized target organ $o$, a set of anatomy-level target locations $a$, and the supporting
textual basis $b$. This representation was designed to preserve both clinical traceability and
anatomical usefulness for downstream grounding.

A staged LLM-based distillation pipeline was used to derive these units from the raw clinical text.
In the first stage, descriptive clinical content was extracted over the full report, including
patient symptoms, examination findings, laboratory or imaging observations, and observable clinical
signs, while explicitly excluding disease labels, diagnostic conclusions, and treatment decisions.
This separation was introduced to reduce diagnosis leakage into the symptom field and to preserve a
cleaner mapping from presenting evidence to downstream anatomical grounding. In the second stage,
target organs were extracted only from physician-authored diagnostic or assessment sections, such
as impression, diagnosis, assessment-and-plan, and discharge diagnosis. This diagnosis-anchored
design reduces free-form organ hallucination and ensures that organ assignment remains tied to
explicit medical judgment rather than inferred solely from symptom text. In the third stage, each
symptom--organ pair was mapped to anatomy-level targets. When specific anatomical parts were
explicitly mentioned in the source text, they were preserved directly; otherwise, relevant target
structures were selected under a predefined organ--anatomy ontology. Organ names were normalized to
a fixed whitelist of standard anatomical entities (that is, a fixed allowed target set), and
anatomy-level targets were restricted to
predefined structure sets for supported organs. Distilled outputs were then normalized into a
unified schema, validated for structural consistency, and filtered by organ-name and
anatomy-location constraints before insertion into the semantic prior database.

After distillation, all semantic units were stored as retrieval records indexed by symptom/sign
text, diagnosis anchor, target organ, anatomy-level locations, and supporting textual evidence. At
inference time, a clinical query or symptom description was used to retrieve semantically relevant
records from the indexed database, and the retrieved evidence was then aggregated to ground the
input into the target tuple $(o,\tau,r)$: the grounded target organ $o$, a prioritized anatomical
region or ROI $r$ resolved from the retrieved anatomy-level locations $a$, and a task type $\tau$
inferred from the query together with the retrieved evidence. Related secondary structures may also
be retained as auxiliary context. Because the database is organized directly at the
symptom--diagnosis--organ--anatomical-location level, retrieval operates in the same semantic space
as the downstream grounding task rather than through unconstrained free-text generation. In this
implementation, prompt templates, text cleaning, chunking, normalization, and indexing
configuration are all treated as part of the semantic-layer implementation details.

\subsection*{Canonical parametric-body space: transform definition and scale handling}\label{supp:canon_body}

In Section~\ref{sec:methods-lowdim}, the canonical coordinate frame is defined as the case-specific MHR rest/T-pose 
rig frame. To obtain the explicit rig geometry required for organ anchoring, 
per-joint global transforms $G_j$ are computed via forward kinematics. Throughout the Methods and
Supplementary Methods, spatial points are written as row vectors, and affine transforms act from the
right on augmented row coordinates $\tilde v = [\,v\;\;1\,]$. In the simplified view, each joint 
transform consists of a rotation $R_j$ and a translation $t_j$. In this implementation, the MHR 
skeleton state additionally provides a per-joint uniform scale $s_j$, which is retained for 
consistency with the canonicalization step:

\begin{equation}
G_j =
\begin{bmatrix}
 (s_j R_j)^\top & 0 \\
 t_j^\top & 1
\end{bmatrix},
\qquad
\tilde v' = \tilde v\, G_j,
\qquad R_j \in \mathrm{SO}(3), \qquad t_j \in \mathbb{R}^3, \qquad s_j > 0.
\end{equation}

Retaining $s_j$ ensures that the same joint transforms are used throughout canonicalization and 
subsequent synthesis, avoiding train--test mismatches caused by dropping the scale term.

\subsection*{Skin registration: optimization details and SMPL-H-to-MHR conversion}\label{supp:skin_reg}

For each subject, skin registration is performed in two stages. A pose-only 
initialization driven by the extracted 3D anatomical keypoints is first run to obtain a stable skeletal 
alignment. Pose and shape are then refined jointly using bidirectional surface distances between the 
observed skin mesh and the SMPL-H surface together with low-dimensional regularization terms. The 
full objective is

\begin{equation}
\begin{aligned}
\mathcal{E} ={}& \lambda_{\mathrm{s2m}}\,\mathcal{E}_{\mathrm{s2m}}(V^{\mathrm{SMPLH}}_i, V^{\mathrm{skin}}_i)
+ \lambda_{\mathrm{m2s}}\,\mathcal{E}_{\mathrm{m2s}}(V^{\mathrm{SMPLH}}_i, V^{\mathrm{skin}}_i) \\
&+ \lambda_{\mathrm{j3d}} \sum_m w_m \left\| J_m(\beta_i, \theta_i, t_i) - p_m \right\|_2^2
+ \lambda_{\theta}\,\mathcal{E}_{\mathrm{pose\mbox{-}prior}}(\theta_i) \\
&+ \lambda_{\beta}\,\lVert \beta_i \rVert_2^2
+ \lambda_{\mathrm{hand}}\,\mathcal{E}_{\mathrm{hand\mbox{-}prior}}.
\end{aligned}
\end{equation}

Here, $\mathcal{E}_{\mathrm{s2m}}$ and $\mathcal{E}_{\mathrm{m2s}}$ are the scan-to-model and 
model-to-scan surface distance terms, respectively; $J_m(\cdot)$ is the SMPL-H joint function for 
observed keypoint $m$, with $p_m$ denoting the corresponding observed 3D keypoint; and $\mathcal{E}_{\mathrm{pose\mbox{-}prior}}$ and 
$\mathcal{E}_{\mathrm{hand\mbox{-}prior}}$ regularize body and hand articulation. The optimization 
outputs an SMPL-H parameter package $\{\beta_i, \theta_i, t_i\}$.

The fitted body state is then converted to the MHR model following the official MHR pipeline 
\cite{MHR:2025}. Rather than relying on a hand-crafted parameter mapping between SMPL-H and MHR, 
this conversion proceeds by: (i) generating the posed SMPL-H surface from the fitted parameters, 
(ii) transferring that surface to the MHR topology via barycentric interpolation, and 
(iii) optimizing MHR parameters to match the transferred target surface. During this conversion, 
unit consistency is handled explicitly (SMPL-H in meters and MHR in centimeters), yielding an MHR 
parameter set together with the corresponding MHR mesh and rig state used in the subsequent 
canonicalization stages.

\subsection*{Canonicalization via organ-aware inverse skinning: weight assignment, scale handling, and regularization}\label{supp:canon_invskin}

After converting each fitted body to the MHR rig, organ meshes are unposed from the posed 
configuration back to the subject-specific canonical rest-rig space using an approximate inverse skinning 
operation. In this step, skinning weights are assigned from the local joint geometry using 
inverse-squared distances to candidate joints. Here, $d(v,j)$ denotes the Euclidean distance
between vertex $v$ and joint $j$, $\epsilon>0$ is a small constant for numerical stability, and
$\mathcal{N}(v)$ is the local candidate-joint set assigned to vertex $v$:

\begin{equation}
w_{v,j} = \frac{\left(d(v,j)^2 + \epsilon\right)^{-1}}
{\sum_{k \in \mathcal{N}(v)} \left(d(v,k)^2 + \epsilon\right)^{-1}}.
\end{equation}

This weighting provides smooth spatial blending and reduces deformation artifacts associated with 
hard joint assignments. To improve stability for internal organs, an organ-aware nested 
constraint is applied: candidate joints are first restricted to a trunk-related whitelist (a fixed
allowed joint set comprising spine, root, hips, 
and neck), and then further restricted to a small organ-centric support that is determined from the organ centroid. In this way, each organ is canonicalized using a compact and semantically relevant 
subset of joints rather than arbitrary body-wide support. In notation, this step maps
$v^{\mathrm{pose}} \mapsto v^{\mathrm{canon}}$ by blending the right-acting joint-wise unposing transforms
$(G_{i,j}^{\mathrm{pose}})^{-1}G_{i,j}^{\mathrm{rest}}$; it is therefore an approximate inverse
skinning operation, not the exact inverse of the blended forward linear blend skinning (LBS) map.

As described in Section~\ref{sec:methods-lowdim}, the MHR skeleton state includes a per-joint uniform scale term 
$s_j$. This scale is retained in the homogeneous transforms $G_{i,j}$ for both posed and rest rigs 
so that the same transform convention is used throughout canonicalization and subsequent synthesis. 
This avoids scale-related mismatches between offline processing and online instantiation.

Finally, canonicalized assets are represented in centimeters to align with the MHR canonical
rest-rig space 
representation, with explicit unit conversion applied where intermediate stages use meters. Light
geometric regularization, including isotropic smoothing and boundary stitching, is also applied to 
$V_i^{\mathrm{canon}}$ in order to improve surface regularity before the subsequent dense 
correspondence step. The geometry-derived 3D keypoints used in this registration stage are listed
in Table~\ref{tab:keypoints_geometry}.

\begin{table}[t]
\centering
\small
\setlength{\tabcolsep}{4pt}
\renewcommand{\arraystretch}{1.10}
\caption{Geometry-derived 3D keypoints from segmented skeletal meshes.}
\label{tab:keypoints_geometry}%
\begin{tabular}{@{}p{0.50\linewidth}p{0.42\linewidth}@{}}
\toprule
\textbf{Keypoint(s)} & \textbf{Source mesh(es)} \\
\midrule
RHip, LHip & Femur (R/L) \\
MidHip & RHip/LHip, pelvis \\
RShoulder, LShoulder & Humerus (R/L) \\
RElbow, LElbow & Humerus (R/L) \\
RWrist, LWrist & Ulna (R/L) \\
RKnee, LKnee & Femur (R/L) \\
RAnkle, LAnkle & Tibia (R/L) \\
RHeel, LHeel & Tarsal (R/L) \\
Big toe / small toe (R/L) & Toes (R/L) \\
Neck & Spine \\
Nose & Skull \\
Eyes, Ears & Skull \\
\bottomrule
\end{tabular}
\end{table}

\subsection*{Dense correspondence in canonical space via template registration}\label{supp:dense_corr}

Following the canonicalization step summarized in Section~\ref{sec:methods-canonical-construction}, each subject provides an unposed
organ mesh in a subject-specific rest-rig space. However, because these raw meshes originate from
independent patient-specific extractions, they lack the consistent topology and vertex indexing
required for population-level statistical modeling. To resolve this, dense
correspondence is established by non-rigidly registering a semantic organ template to each subject's unposed
target surface in canonical space.

For a given organ $k$, a canonical-space template mesh $V_k$ is used as the source. This template
is an anatomically curated model designed to preserve internal geometric semantics, with a moderate
resolution of approximately 7{,}000--10{,}000 faces to balance geometric fidelity and computational
efficiency. The registration process deforms the template vertices while strictly preserving the
original face connectivity. This deformation is parameterized using a global translation vector
$t_{i,k}^{\mathrm{reg}}$ and a per-vertex spatial offset field $\Delta V_{i,k}$:
\begin{equation}
\hat{V}_{i,k} = V_k + \Delta V_{i,k} + \mathbf{1}_{n_k} (t_{i,k}^{\mathrm{reg}})^{\top},
\end{equation}
where $\mathbf{1}_{n_k}\in\mathbb{R}^{n_k\times 1}$ denotes the all-ones column vector for organ
$k$ with $n_k$ template vertices. Because the preceding rig-anchored canonicalization step already
resolves macroscopic pose-induced spatial discrepancies, the target meshes are well aligned in
canonical space, allowing the non-rigid registration to converge robustly from translational
initialization.

To ensure physical plausibility and geometric fidelity, the deformation is estimated by minimizing
the composite objective
\begin{equation}
E = \lambda_{\mathrm{data}} E_{\mathrm{chamfer}} + \lambda_{\mathrm{edge}} E_{\mathrm{edge}} +
\lambda_{\mathrm{normal}} E_{\mathrm{normal}} + \lambda_{\mathrm{lap}} E_{\mathrm{lap}}.
\end{equation}
Here, the data term $E_{\mathrm{chamfer}}$ uses the Chamfer distance between point samples drawn
from the deformed template and the target organ surface to enforce bidirectional geometric
proximity. The remaining terms regularize the offset field: $E_{\mathrm{edge}}$ limits local
stretching, $E_{\mathrm{normal}}$ discourages fold-overs and normal flipping, and
$E_{\mathrm{lap}}$ enforces Laplacian smoothness. In this implementation, optimization follows a
three-stage curriculum: coarse translation initialization, joint optimization of translation and
local offsets, and final smoothing refinement with stronger regularization to suppress
high-frequency artifacts.

Because the optimizer updates only the vertex coordinates while keeping the template connectivity
immutable, each resulting registered mesh $T_{i,k}$ is a morphologically adapted instance of the
same template $V_k$. Consequently, the full registered set $\{T_{i,k}\}_{i=1}^{N}$ shares a
uniform topology and vertex-index semantics across subjects, providing the correspondence basis for
the low-dimensional representation and prior learning stages in Sections~\ref{sec:methods-lowdim}--\ref{sec:methods-skeleton-prior}.

\subsection*{Skeleton-conditioned prior learning: whitelist, feature construction, regression, and saved assets}\label{supp:skeleton_prior}

This subsection provides the implementation details omitted from the condensed presentation in
Section~\ref{sec:methods-skeleton-prior}. Given the low-dimensional organ representation defined in Section~\ref{sec:methods-lowdim}, prior
learning is implemented as organ-wise regression on the supervised targets $\Delta c_{i,k}$ and
$\ell_{i,k}$, while the instance-specific rotations are aggregated into an organ-wise mean
orientation prior $\bar R_k$.

The regression features are extracted from the subject-specific rest rig rather than from posed
skeletal states. To ensure anatomical relevance and train--test consistency, joint selection is
performed under a whitelist-based rule. The candidate pool is first restricted to a trunk-related
skeletal whitelist, comprising spine- and root-anchored segments together with adjacent pelvic and
upper-trunk joints, and a small organ-specific subset is then defined within that pool. In the
default automatic configuration, this subset is selected on a reference sample by organ-centroid
proximity. Specifically, if no manual subset is provided, the implementation selects the nearest
$N$ joints to the reference organ centroid, with $N=3$ in the default setting. The same
whitelist-defined joint subset is then reused unchanged for all training and test cases of that
organ.

Let $J_{i,k}^{\mathrm{local}}$ denote the selected rest-rig joints expressed in the subject's
anatomical local frame. From these joints, an explicit low-dimensional geometric
feature vector is constructed:
\begin{equation}
x_{i,k} = \phi\left(J_{i,k}^{\mathrm{local}}\right).
\end{equation}
In the current implementation, $\phi(\cdot)$ consists of three components: (i) relative joint
positions, defined as the 3D offsets of the remaining selected joints with respect to the first
joint; (ii) pairwise Euclidean distances between all selected joint pairs; and (iii) a single
angle-cosine term computed from the first three selected joints when available. Under the default
configuration with three selected joints and with both distance and angle terms enabled, this
yields a 10-dimensional feature vector comprising two 3D relative offsets, three pairwise
distances, and one angle cosine. When enabled, the first $K_\beta$ body-shape coefficients are
concatenated to the skeletal feature vector as an optional residual correction term,
\begin{equation}
\tilde x_{i,k} = [x_{i,k}; \beta_i^{1:K_\beta}],
\end{equation}
where $K_\beta$ denotes the number of retained shape coefficients; in the default experiments,
skeleton-only features are used without $\beta$ augmentation.

For each organ class, two organ-wise regression models are fit,
\begin{equation}
\hat{\Delta c}_{i,k} = f_{\mathrm{pos},k}(\tilde x_{i,k}),
\qquad
\widehat{\ell}_{i,k} = f_{\mathrm{scale},k}(\tilde x_{i,k}).
\end{equation}
In the default implementation, both $f_{\mathrm{pos},k}$ and $f_{\mathrm{scale},k}$ are linear
mappings of the form
\begin{equation}
f(\tilde x) = W\tilde x + b,
\end{equation}
implemented as ordinary least-squares regression. An optional ridge-regularized variant is also
supported for robustness, in which the parameters are estimated by minimizing
\begin{equation}
\min_{W,b} \sum_i \left\| y_{i,k} - (W\tilde x_{i,k} + b) \right\|_2^2 + \lambda \|W\|_F^2,
\end{equation}
where $y_{i,k}$ denotes the supervised target for one regression head of organ class $k$, i.e.,
either $\Delta c_{i,k}$ (position) or $\ell_{i,k}$ (scale). Rotation is not learned as a
subject-conditioned predictor in the current implementation. Instead, the instance-specific
rotations extracted during decomposition are aggregated over the training set to estimate an
organ-wise mean orientation prior $\bar R_k$ for
reconstruction.

Before fitting the organ-wise regressors, abnormal samples are removed at the level of the
low-dimensional geometric decomposition. Outlier filtering is applied to two quantities derived from
Section~\ref{sec:methods-lowdim}: the rigid-alignment RMSE and the organ volume ratio, where the latter is computed
from the anisotropic scale factors as the product of the three axis-wise scales. For each quantity,
samples outside the interquartile-range bounds are rejected using the default multiplier of 1.5,
and the resulting keep-mask is applied consistently to the decomposed supervision targets and to
the corresponding skeletal feature matrix. This step removes unstable decompositions before
regression and improves robustness in the small-sample regime.

The current implementation models uncertainty at the level of regression residuals rather than as a
full posterior over organ geometry. After fitting the position and scale regressors, axis-wise residual variances are estimated
for centroid displacement and logarithmic scale, respectively:
\begin{equation}
\boldsymbol{\sigma}^2_{\mathrm{pos},k} =
\mathrm{Var}_{\mathrm{axis}}\!\left(\Delta c_{i,k} - \hat{\Delta c}_{i,k}\right) \in \mathbb{R}^3,
\qquad
\boldsymbol{\sigma}^2_{\mathrm{scale},k} =
\mathrm{Var}_{\mathrm{axis}}\!\left(\ell_{i,k} - \widehat{\ell}_{i,k}\right) \in \mathbb{R}^3,
\end{equation}
where both quantities are stored as 3D vectors corresponding to the three anatomical axes. In
addition, the empirical covariance of the decomposed centroid displacements is retained,
\begin{equation}
\Sigma_{\Delta c,k} = \mathrm{Cov}(\Delta c_{i,k}),
\end{equation}
which summarizes the population-level spread of organ placement in the anatomical local frame.

To ensure train--test consistency, the training stage saves not only the fitted position and scale
regressors, but also the whitelist-defined joint subset, the feature-construction configuration,
the organ-wise mean orientation prior, and the reference coordinate-system instance used to preserve
the template-local frame. At inference time, the same fixed joint indices and the same feature
definition are reapplied to the new subject's rest rig, optionally with the same $\beta$
concatenation rule if enabled during training. The predicted outputs are therefore
\begin{equation}
\hat{\Delta c}_{i,k},
\qquad
\widehat{\ell}_{i,k},
\qquad
\hat s_{i,k} = \exp(\widehat{\ell}_{i,k}),
\end{equation}
while rotation is provided by the stored organ-wise mean rotation. Together with the uncertainty
statistics above, these saved assets define the complete offline prior used in the online
instantiation stage.

\subsection*{Additional organ-wise evaluation on Quadra-HC}\label{supp:ari_quadrahc}

For completeness, Supplementary Table~\ref{tab:supp_ari_organs} reports the organ-wise
quantitative results of Anatomical Representation Instantiation on the Quadra-HC evaluation set.
Each row summarizes one organ over 35 held-out cases.

\begin{table}[t]
\centering
\footnotesize
\setlength{\tabcolsep}{0pt}
\renewcommand{\arraystretch}{1.12}
\begin{tabular*}{\linewidth}{@{\extracolsep{\fill}}lccccccc@{}}
\toprule
Organ & N & \shortstack{Centroid\\err. (mm)} & \multicolumn{3}{c}{\shortstack{Per-axis centroid err.\\(mm)}} & \shortstack{Scale\\err. (\%)} & \shortstack{Support\\IoU} \\
\cmidrule(lr){4-6}
 &  &  & x & y & z &  &  \\
\midrule
heart & 35 & 19.90 & 10.92 & 8.74 & 10.91 & 11.13 & 0.567 \\
liver & 35 & 26.48 & 10.17 & 9.88 & 19.66 & 18.79 & 0.472 \\
spleen & 35 & 23.47 & 11.41 & 11.85 & 12.20 & 15.98 & 0.421 \\
pancreas & 35 & 23.07 & 13.73 & 6.56 & 13.60 & 24.65 & 0.303 \\
bladder & 35 & 26.41 & 8.05 & 8.07 & 21.47 & 21.91 & 0.226 \\
aorta & 35 & 16.52 & 7.08 & 3.64 & 11.94 & 14.67 & 0.369 \\
thyroid & 35 & 19.41 & 6.67 & 11.52 & 10.22 & 14.12 & 0.186 \\
inferior vena cava & 35 & 25.39 & 7.28 & 9.82 & 19.56 & 18.41 & 0.338 \\
lung & 35 & 11.99 & 4.64 & 4.35 & 8.96 & 10.35 & 0.678 \\
left kidney & 35 & 30.61 & 14.09 & 13.75 & 19.35 & 14.09 & 0.327 \\
right kidney & 35 & 24.82 & 5.19 & 13.30 & 18.09 & 12.96 & 0.413 \\
\bottomrule
\end{tabular*}
\caption{Organ-wise Anatomical Representation Instantiation results on the Quadra-HC evaluation
set. Entries are shown as mean values over 35 held-out cases for each organ. Centroid error
denotes mean Euclidean centroid error. The x, y, and z columns denote per-axis mean absolute
centroid error. Scale error denotes bounding-box extent relative error, and support extent
consistency denotes bounding-box-support IoU.}
\label{tab:supp_ari_organs}
\end{table}

\subsection*{Additional ATI ablation results}\label{supp:ati_ablation}

For completeness, Supplementary Table~\ref{tab:supp_ati_ablation} reports the paired ablation used
to examine the contribution of Actionable Target Initialization for centroid targets. The full
SAMe stack was compared with an ARI-only variant using the same instantiated anatomical targets.
In the ARI-only variant, the contact-aware ATI stage was removed.

\begin{table}[t]
\centering
\small
\setlength{\tabcolsep}{5pt}
\renewcommand{\arraystretch}{1.05}
\begin{tabular}{@{}llcccc@{}}
\toprule
& & \multicolumn{2}{c}{Trial-level organ-hit} & \multicolumn{2}{c}{Trial-level anatomy match} \\
\cmidrule(lr){3-4}\cmidrule(lr){5-6}
Organ & Target & w/ ATI & w/o ATI & w/ ATI & w/o ATI \\
\midrule
\multirow{1}{*}{Kidney}
& Centroid & 13/15 (87\%) & 12/15 (80\%) & 12/15 (80\%) & 10/15 (67\%) \\
\midrule
\multirow{1}{*}{Liver}
& Centroid & 13/15 (87\%) & 12/15 (80\%) & 12/15 (80\%) & 11/15 (73\%) \\
\bottomrule
\end{tabular}
\caption{Paired ablation of Actionable Target Initialization. Entries report successful scans over
valid trials for organ-level hit and anatomy-level target match for centroid targets.}
\label{tab:supp_ati_ablation}
\end{table}

\subsection*{Vector definitions for body-surface projection and contact candidate ranking}\label{supp:vector_defs}

Let $p_{\mathrm{tar}} \in \mathbb{R}^3$ denote the internal target point, which may correspond 
to an organ centroid, an ROI center, or an anatomical landmark. Let the subject-specific 
anatomical coordinate system be given by origin $o^{\mathrm{ACS}}$ and orthonormal axes 
$(\mathbf{e}_x, \mathbf{e}_y, \mathbf{e}_z)$, where $\mathbf{e}_z$ denotes the anterior body 
direction used in the current implementation. The forward projection ray is defined as
\begin{equation}
\mathbf{r}_{\mathrm{proj}}(\lambda) = p_{\mathrm{tar}} + \lambda \, \mathbf{e}_z,
\qquad \lambda \ge 0.
\end{equation}
The initial body-surface projection point $p_{\mathrm{proj}}$ is obtained as the first 
intersection of this ray with the skin mesh $\mathcal{M}_{\mathrm{skin}}$:
\begin{equation}
p_{\mathrm{proj}} = \operatorname*{arg\,min}_{\lambda \ge 0}
\left\{ \mathbf{r}_{\mathrm{proj}}(\lambda) \in \mathcal{M}_{\mathrm{skin}} \right\}.
\end{equation}

To refine this initialization, a local forward-facing hemispherical candidate set is defined on the 
skin surface around $p_{\mathrm{proj}}$. Let $q \in \mathcal{M}_{\mathrm{skin}}$ be a candidate 
surface point in this neighborhood, and let $\mathbf{n}(q)$ be its outward unit surface normal. 
The target-directed unit ray from $q$ toward the internal target is defined as
\begin{equation}
\mathbf{r}(q) = \frac{p_{\mathrm{tar}} - q}{\left\| p_{\mathrm{tar}} - q \right\|_2}.
\end{equation}
Normal alignment is then measured by the opposition between the outward surface normal and the
target-directed ray:
\begin{equation}
s_{\mathrm{align}}(q) = -\,\mathbf{n}(q)^\top \mathbf{r}(q).
\end{equation}
A larger value of $s_{\mathrm{align}}(q)$ indicates that the outward normal points more directly
opposite to the direction from the skin toward the internal target, and is therefore preferred for
initial ultrasound access. A second component penalizes candidates whose probe entry path intersects
skeletal structures, yielding a skeletal clearance term $s_{\mathrm{skel}}(q) \in [0, 1]$. Let
$\mathcal{L}(q)$ denote the line segment from $q$ to $p_{\mathrm{tar}}$, let
$\mathcal{M}_{\mathrm{skel}}$ denote the predicted skeletal surface, and define
\begin{equation}
d_{\mathrm{skel}}(q)=\min_{x \in \mathcal{L}(q),\, y \in \mathcal{M}_{\mathrm{skel}}}\|x-y\|_2,
\qquad
s_{\mathrm{skel}}(q)=\min\!\left(1,\frac{d_{\mathrm{skel}}(q)}{\delta_{\mathrm{skel}}}\right),
\end{equation}
where $\delta_{\mathrm{skel}} > 0$ is a fixed clearance threshold. The composite score is
\begin{equation}
s(q) = s_{\mathrm{align}}(q) \cdot s_{\mathrm{skel}}(q).
\end{equation}
When bone-level prediction is unavailable, $s_{\mathrm{skel}}(q) = 1$ for all candidates and the
scoring reduces to geometric alignment alone. In the current implementation, candidate points are
ranked by $s(q)$, and the top $K_{\mathrm{cand}}$ candidates (with $K_{\mathrm{cand}}=3$ in the
present experiments) are retained:
\begin{equation}
\mathcal{Q}_{\mathrm{top}} = \operatorname*{TopK}_{q \in \mathcal{Q},\,K_{\mathrm{cand}}} \, s(q),
\end{equation}
where $\mathcal{Q}$ denotes the local hemispherical candidate set around $p_{\mathrm{proj}}$.

For each selected candidate $q \in \mathcal{Q}_{\mathrm{top}}$, the downstream control-facing 
geometric quantities consist of
\begin{equation}
\bigl(q,\; \mathbf{n}(q),\; \mathbf{r}(q)\bigr),
\end{equation}
namely the candidate surface contact point, its associated skin normal, and the target-directed 
entry ray. These quantities define an initialization-ready geometric constraint set for 
downstream probe placement and contact-aware orientation setup.

\subsection*{Notation summary}\label{supp:notation}

This subsection consolidates the notation used throughout the main
text and Supplementary Methods.
Indices, coordinate frames, learned variables, and control-facing
geometric quantities are listed
separately to keep the symbol system explicit and consistent across
the manuscript. Unless stated
otherwise, subject/case indices are denoted by $i$, organ-class
indices by $k$, joint indices by
$j$, observed keypoint indices by $m$, and vertex symbols by $v$.

{\small
\renewcommand{\arraystretch}{1.2}
\setlength{\LTleft}{0pt}
\setlength{\LTright}{0pt}
\setlength{\LTpre}{4pt}
\setlength{\LTpost}{3pt}

\begin{longtable}{@{}L{0.12\linewidth}L{0.82\linewidth}@{}}
\toprule
\textbf{Symbol} & \textbf{Meaning} \\
\midrule
\label{tab:notation_indices}
\endfirsthead
\multicolumn{2}{c}{\small\textit{(continued)}} \\
\toprule
\textbf{Symbol} & \textbf{Meaning} \\
\midrule
\endhead
\midrule
\multicolumn{2}{r}{\small\textit{(continued on next page)}} \\
\endfoot
\bottomrule
\endlastfoot

$i$ & Subject or case index \\
$k$ & Organ-class index \\
$j$ & Joint index \\
$m$ & Observed anatomical keypoint index \\
$v$ & Mesh vertex \\
$q$ & Candidate skin-surface contact point \\
$\lambda$ & Scalar ray parameter \\
$c(\cdot)$ & Centroid operator applied to a mesh or point set \\
$\phi(\cdot)$ & Explicit skeletal feature extractor \\
$\psi_k(\cdot)$ & Organ-wise mapping from subject-specific body cues to organ placement/scale descriptors \\
$\Sigma$ & Covariance matrix \\
$\boldsymbol{\sigma}^2$ & Axis-wise variance vector \\
$\epsilon$ & Small constant used for numerical stability \\
\end{longtable}
\noindent Table~\thetable: Indices, operators, and generic symbols used throughout the manuscript.
\par\smallskip

\begin{longtable}{@{}L{0.15\linewidth}L{0.79\linewidth}@{}}
\toprule
\textbf{Symbol} & \textbf{Meaning} \\
\midrule
\label{tab:notation_system}
\endfirsthead
\multicolumn{2}{c}{\small\textit{(continued)}} \\
\toprule
\textbf{Symbol} & \textbf{Meaning} \\
\midrule
\endhead
\midrule
\multicolumn{2}{r}{\small\textit{(continued on next page)}} \\
\endfoot
\bottomrule
\endlastfoot

$c$ & Clinical query, complaint description, or examination intent \\
$I$ & Input monocular RGB body image \\
$x_b$ & Body observations derived from external body sensing, including skeletal, shape, and surface cues \\
$x_s \subseteq x_b$ & Subset of body observations used specifically for body-surface initialization \\
$o$ & Grounded target organ \\
$r$ & Prioritized anatomical region or region of interest (ROI) \\
$\tau$ & Task type inferred during semantic grounding \\
$u=(s,d,o,a,b)$ & Structured semantic unit used in the semantic-prior database \\
$s$ & Symptom or sign description in a semantic unit \\
$d$ & Diagnosis anchor in a semantic unit \\
$a$ & Anatomy-level target location set in a semantic unit \\
$b$ & Supporting textual basis in a semantic unit \\
$z_a$ & Patient-specific anatomical representation produced by Anatomical Representation Instantiation \\
$z_p$ & Abstract initialization output produced by Actionable Target Initialization \\
$\zeta_k$ & Organ-specific control-facing geometric state used as the implementation-level realization of $z_p$ \\
\end{longtable}
\noindent Table~\thetable: System-level inputs, semantic variables, and stage-wise outputs.
\par\smallskip

\begin{longtable}{@{}L{0.22\linewidth}L{0.74\linewidth}@{}}
\toprule
\textbf{Symbol} & \textbf{Meaning} \\
\midrule
\label{tab:notation_geometry}
\endfirsthead
\multicolumn{2}{c}{\small\textit{(continued)}} \\
\toprule
\textbf{Symbol} & \textbf{Meaning} \\
\midrule
\endhead
\midrule
\multicolumn{2}{r}{\small\textit{(continued on next page)}} \\
\endfoot
\bottomrule
\endlastfoot

$M=(V,F)$ & Generic triangular mesh with vertex set $V$ and face set $F$ \\
$\mathcal{A}_i$ & Mesh asset set of subject $i$ \\
$S_i$ & Skin surface set of subject $i$ \\
$B_i$ & Skeletal surface set of subject $i$ \\
$O_i^c$ & Core organ set of subject $i$ \\
$O_i^a$ & Auxiliary organ set of subject $i$ \\
$M_i^{\mathrm{skin}}$ & Skin mesh of subject $i$ \\
$M_{i,k}^{\mathrm{organ}}$ & Organ mesh of subject $i$ for organ class $k$ \\
$G_j$ & Generic per-joint homogeneous transform acting on augmented row coordinates \\
$R_j$ & Joint rotation component \\
$t_j$ & Joint translation component \\
$s_j$ & Per-joint uniform scale \\
$G_{i,j}^{\mathrm{rest}}$ & Rest-rig transform of joint $j$ for subject $i$ \\
$G_{i,j}^{\mathrm{pose}}$ & Posed-rig transform of joint $j$ for subject $i$ \\
$\Delta G_{i,j}$ & Joint-wise unposing transform, defined as $(G_{i,j}^{\mathrm{pose}})^{-1}G_{i,j}^{\mathrm{rest}}$ \\
$o_i^{\mathrm{ACS}}$ & Origin of the anatomical coordinate system (ACS) for subject $i$ \\
$R_i^{\mathrm{ACS}}$ & Orientation matrix of the ACS for subject $i$ \\
$V_{i,k}^{\mathrm{world}}$ & Registered organ mesh of subject $i$, organ $k$, in world coordinates \\
$V_{i,k}^{\mathrm{local}}$ & Registered organ mesh of subject $i$, organ $k$, in the subject-specific ACS \\
$V_k$ & Canonical organ template mesh for organ class $k$ \\
$V_{k,\mathrm{temp}}^{\mathrm{world}}$ & Reference template mesh for organ class $k$ in world coordinates \\
$V_{k,\mathrm{temp}}^{\mathrm{local}}$ & Reference template mesh for organ class $k$ in the reference local frame \\
$o_{\mathrm{ref}}^{\mathrm{ACS}}$ & ACS origin of the selected reference case \\
$R_{\mathrm{ref}}^{\mathrm{ACS}}$ & ACS orientation of the selected reference case \\
$v^{\mathrm{pose}}$ & Vertex in posed/acquisition space \\
$v^{\mathrm{canon}}$ & Canonicalized vertex in rest-rig space \\
$\mathcal{N}(v)$ & Candidate-joint set used for vertex $v$ in inverse skinning \\
$d(v,j)$ & Euclidean distance between vertex $v$ and joint $j$ \\
$w_{v,j}$ & Inverse-skinning blending weight of joint $j$ for vertex $v$ \\
\end{longtable}
\noindent Table~\thetable: Mesh, coordinate-frame, and transformation notation.
\par\smallskip

\begin{longtable}{@{}L{0.22\linewidth}L{0.74\linewidth}@{}}
\toprule
\textbf{Symbol} & \textbf{Meaning} \\
\midrule
\label{tab:notation_prior}
\endfirsthead
\multicolumn{2}{c}{\small\textit{(continued)}} \\
\toprule
\textbf{Symbol} & \textbf{Meaning} \\
\midrule
\endhead
\midrule
\multicolumn{2}{r}{\small\textit{(continued on next page)}} \\
\endfoot
\bottomrule
\endlastfoot

$\mathcal{E}$ & Overall optimization objective in the corresponding stage \\
$\mathcal{E}_{\mathrm{s2m}}$ & Scan-to-model surface alignment term \\
$\mathcal{E}_{\mathrm{m2s}}$ & Model-to-scan surface alignment term \\
$V_i^{\mathrm{SMPLH}}$ & SMPL-H surface of subject $i$ \\
$V_i^{\mathrm{skin}}$ & Observed skin surface of subject $i$ \\
$\beta_i$ & Body-shape coefficients of subject $i$ \\
$\theta_i$ & Body-pose parameters of subject $i$ \\
$t_i$ & Global translation of subject $i$ in body-model fitting \\
$J_m(\beta_i,\theta_i,t_i)$ & Model joint or landmark function corresponding to observed keypoint $m$ \\
$p_m$ & Observed 3D anatomical keypoint corresponding to observed keypoint index $m$ \\
$\hat{V}_{i,k}$ & Deformed template mesh after dense registration for subject $i$, organ $k$ \\
$\Delta V_{i,k}$ & Per-vertex deformation field for subject $i$, organ $k$ \\
$t_{i,k}^{\mathrm{reg}}$ & Global translation used in dense correspondence registration \\
$\mathbf{1}_{n_k}$ & All-ones column vector with $n_k$ entries \\
$n_k$ & Number of template vertices for organ class $k$ \\
$T_{i,k}$ & Topology-consistent registered organ instance of subject $i$, organ $k$ \\
$R_{i,k}$ & Rigid alignment rotation descriptor for subject $i$, organ $k$ \\
$t_{i,k}^{\mathrm{rigid}}$ & Rigid alignment translation for subject $i$, organ $k$ \\
$\tilde V_{i,k}$ & Rigidly aligned organ instance used for scale estimation \\
$\Delta c_{i,k}$ & Template-relative centroid displacement of organ $k$ in subject $i$ \\
$s_{i,k}$ & Anisotropic scale of organ $k$ in subject $i$ \\
$\ell_{i,k}$ & Log-anisotropic scale of organ $k$ in subject $i$, defined as $\ell_{i,k}=\log s_{i,k}$ \\
$\bar R_k$ & Organ-wise mean orientation prior for organ class $k$ \\
$\bar c_{k,\mathrm{temp}}$ & Centroid of the reference-local template mesh for organ class $k$ \\
$\mathcal{D}_k$ & Training set for organ class $k$ \\
$N_k$ & Number of training samples for organ class $k$ \\
$J_{i,k}^{\mathrm{local}}$ & Selected local skeletal joints used for organ class $k$ in subject $i$ \\
$x_{i,k}$ & Skeletal feature vector extracted for subject $i$, organ $k$ \\
$K_\beta$ & Number of retained shape coefficients used in optional feature augmentation \\
$\tilde x_{i,k}$ & Augmented feature vector formed by concatenating $x_{i,k}$ and optional body-shape coefficients \\
$f_{\mathrm{pos},k}$ & Organ-wise regressor for centroid displacement \\
$f_{\mathrm{scale},k}$ & Organ-wise regressor for logarithmic anisotropic scale \\
$\hat{\Delta c}_{i,k}$ & Predicted centroid displacement for subject $i$, organ $k$ \\
$\widehat{\ell}_{i,k}$ & Predicted logarithmic anisotropic scale for subject $i$, organ $k$ \\
$\hat s_{i,k}$ & Predicted anisotropic scale for subject $i$, organ $k$, defined as $\exp(\widehat{\ell}_{i,k})$ \\
$\boldsymbol{\sigma}_{\mathrm{pos},k}^2$ & Axis-wise residual variance of centroid displacement for organ class $k$ \\
$\boldsymbol{\sigma}_{\mathrm{scale},k}^2$ & Axis-wise residual variance of logarithmic scale for organ class $k$ \\
$\Sigma_{\Delta c,k}$ & Empirical covariance of centroid displacement for organ class $k$ \\
$V_{i,k}^{\mathrm{pred,local}}$ & Predicted organ mesh of subject $i$, organ $k$, in local ACS coordinates \\
$V_{i,k}^{\mathrm{pred,world}}$ & Predicted organ mesh of subject $i$, organ $k$, in world coordinates \\
\end{longtable}
\noindent Table~\thetable: Registration, low-dimensional decomposition, and organ-wise prior learning notation.
\par\smallskip

\begin{longtable}{@{}L{0.22\linewidth}L{0.74\linewidth}@{}}
\toprule
\textbf{Symbol} & \textbf{Meaning} \\
\midrule
\label{tab:notation_ati}
\endfirsthead
\multicolumn{2}{c}{\small\textit{(continued)}} \\
\toprule
\textbf{Symbol} & \textbf{Meaning} \\
\midrule
\endhead
\midrule
\multicolumn{2}{r}{\small\textit{(continued on next page)}} \\
\endfoot
\bottomrule
\endlastfoot

$p_{\mathrm{tar}}$ & Internal target point, such as an organ centroid, ROI center, or anatomical landmark \\
$\mathbf{e}_x,\mathbf{e}_y,\mathbf{e}_z$ & Orthogonal basis vectors of the anatomical coordinate system \\
$\mathbf{e}_z$ & Anterior body direction used for forward body-surface projection \\
$\mathbf{r}_{\mathrm{proj}}(\lambda)$ & Forward projection ray from the internal target toward the body surface \\
$p_{\mathrm{proj}}$ & Initial body-surface projection point obtained from the first ray-surface intersection \\
$\mathcal{M}_{\mathrm{skin}}$ & Skin mesh used for body-surface projection and contact search \\
$\mathcal{Q}$ & Local candidate set of skin-surface points around $p_{\mathrm{proj}}$ \\
$\mathbf{n}(q)$ & Outward unit surface normal at candidate point $q$ \\
$\mathbf{r}(q)$ & Unit ray from candidate point $q$ toward the internal target \\
$s_{\mathrm{align}}(q)$ & Normal-ray alignment score measuring surface--target geometric compatibility \\
$s_{\mathrm{skel}}(q)$ & Skeletal clearance term based on line-segment-to-skeleton distance, $s_{\mathrm{skel}}(q) \in [0, 1]$ \\
$s(q)$ & Composite score, defined as $s(q) = s_{\mathrm{align}}(q) \cdot s_{\mathrm{skel}}(q)$ \\
$K_{\mathrm{cand}}$ & Number of retained candidate contact points \\
$\mathcal{Q}_{\mathrm{top}}$ & Top-ranked candidate set after composite scoring \\
$\zeta_k$ & Control-facing geometric state containing predicted organ geometry, selected contact candidates, orientation prior, and uncertainty descriptors \\
\end{longtable}
\noindent Table~\thetable: Control-facing geometric quantities used in Actionable Target Initialization.
} 

\end{document}